\documentclass[journal]{IEEEtran}
\usepackage{cite}

\ifCLASSINFOpdf

\else

\fi

\usepackage{amsmath}
\usepackage{amsfonts}
\usepackage{amssymb}

\usepackage{graphicx}

\usepackage{float}
\usepackage{subcaption}

\usepackage{algorithm}
\usepackage{algorithmicx}
\usepackage{algpseudocode}

\begin{document}
%
\title{A Learning-Based Approach for Lane Departure Warning Systems with a Personalized \\Driver Model}
\author{Wenshuo~Wang,~
	Ding~Zhao,
	Junqiang~Xi,
	and Wei Han
	\thanks{ The first two authors, W. Wang and D. Zhao, have made equal contributions to this work.}
	\thanks{W. Wang is with the School of Mechanical Engineering, Beijing Institute of Technology (BIT), Beijing, China, 100081, and is studying in Vehicle Dynamics \& Control Lab, Department of Mechanical Engineering, University of California at Berkeley (UCB), CA, 94706 USA. e-mail: wwsbit@gmail.com. wwsvdc2015@berkeley.edu}
	\thanks{D. Zhao is with the University of Michigan Transportation Research Institute, Ann Arbor, MI 48109-2150, USA. e-mail: zhaoding@umich.edu.}
	\thanks{J. Xi is with the School of Mechanical Engineering, Beijing Institute of Technology (BIT), Beijing, China, 100081. e-mail: xijunqiang@bit.edu.cn.}%
	\thanks{W. Han is with the Department of Computer Science, Tsinghua University, Beijing, China, 100084. e-mail: hanwei1949@gmail.com.}
	}


\maketitle

\begin{abstract}
Misunderstanding of  driver correction behaviors (DCB) is the primary reason for false warnings of lane-departure-prediction  systems. We propose a learning-based approach to predicting unintended lane-departure behaviors (LDB) and the chance for drivers to bring the vehicle back to the lane.  First, in this approach, a personalized driver model for lane-departure and lane-keeping behavior is established by combining the Gaussian mixture model and the hidden Markov model. Second, based on this model, we develop an online model-based prediction algorithm to predict the forthcoming vehicle trajectory and judge whether the driver will demonstrate an LDB or a DCB. We also develop a warning strategy based on the model-based prediction algorithm that allows the lane-departure warning system to be acceptable for drivers according to the predicted trajectory. In addition, the naturalistic driving data of 10 drivers is collected through the University of Michigan Safety Pilot Model Deployment program to train the personalized driver model and validate this approach. We compare the proposed method with a basic time-to-lane-crossing (TLC) method and a TLC-directional sequence of piecewise lateral slopes (TLC-DSPLS) method.  The results show that the proposed approach can reduce the false-warning rate to 3.07\%.
%
%
%
%
%
%
%
%
%
%
%
\end{abstract}
\textit{}

\IEEEpeerreviewmaketitle

\section{Introduction}

\subsection{Motivation}

\begin{figure}[t]
	\centering
	\includegraphics[scale = 0.65]{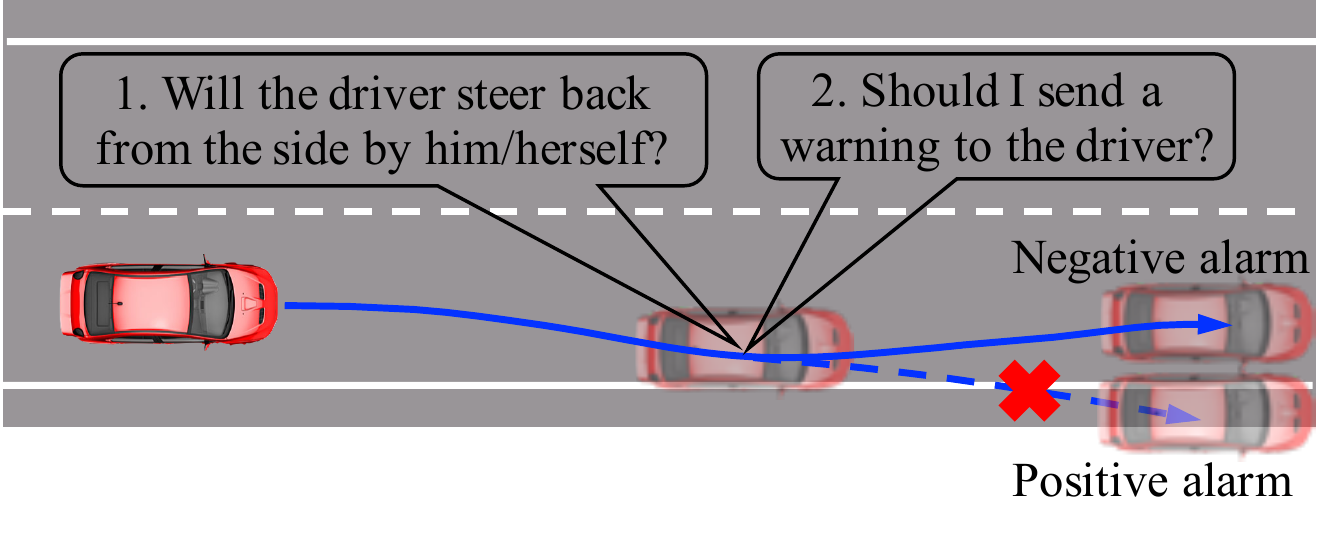}
	\caption{Negative alarm and positive alarm for a lane departure scenario. If the future vehicle trajectory is the dash line, the warning is a positive alarm; if the future vehicle trajectory is the solid line, the warning is a negative alarm.}
	\label{fig.1}
\end{figure}

\IEEEPARstart{L}{ane} departure is an unintentional drifting towards the boundary of the driving lane, which usually occurs when the driver is drowsy or fatigued\cite{may2009driver}. In the United States, 37.4\% of fatal crashes are due to single-vehicle lane departure\cite{barickman2007lane}, which makes it the  leading cause of fatalities\cite{wang1994single}.  Lane-departure warning (LDW) systems aim to alert the driver when the lane departure begins and has great potential for vehicle safety \cite{aksan2016benefits}. LDW systems can detect or predict lane-departure events and give a warning in an auditory, haptic, or visual form \cite{guo2015multimodal,lee2014lane,enache2009driver,angkititrakul2011use,gaikwad2015lane,saito2016driver,lefevre2015driver,enache2010driver}. The challenge to designing a successful LDW is to reduce the false alarm rate (FAR), which occurs when drivers become aware of the lane departure and plan to correct the maneuvers in the next moment \cite{angkititrakul2011use}. A high-rate of false alarms may reduce the driver's trust in the LDW system or cause the driver to become annoyed. As shown in Fig. \ref{fig.1}, the LDW system needs to accurately predict the driver's intention and  provides warnings only when needed. A successful LDW design needs to understand driving style of a specific person and offer a personalized assistance. 

%
%
\subsection{Related Research}

Most LDW systems use time to lane crossing (TLC) to determine whether to activate the warning. TLC is defined as the time duration available for the driver before lane-boundary crossing. However, the TLC-based method has been criticized for having a high FAR \cite{angkititrakul2011use,albousefi2016two} because of its inability to predict driver's intention. The TLC-based warning could be triggered when the TLC reaches the predefined critical value (usually more than 0.9 s). Yet drivers can usually keep the vehicle close to the lane boundary and then bring the vehicle back to the center of the lane without the warning in the second step, as the solid blue line shows in Fig. \ref{fig.1}, which illustrates driver correction behavior (DCB). A warning given when the driver can presumably keep the vehicle in the lane is likely to cause annoyance. Therefore, it is crucial to \textit{infer} the driver's upcoming behavior and \textit{judge} whether the driver could bring the vehicle back to the lane center and then \textit{decide} when the warning should be given to the driver, thus reducing the annoyance level of excessive false warning and avoiding crashes.

Several approaches have been proposed to reduce the FAR. Angkitrakul \textit{et al}.  \cite{angkititrakul2011use} proposed a stochastic driver model to predict the vehicle trajectory. Experimental results showed that the approach had a 17\%  FAR in detecting an intentional correction in the subsequent 0.5 s. When the prediction time was lengthened to 1.5 s, the FAR exceeded 100\%. Another indicator of whether a warning will be given to the driver is based on whether the driver will bring the car back to the lane center after the vehicle approaches the boundary. If the driver plans to bring the vehicle back with some minor departure (i.e., the solid blue line in Fig. \ref{fig.1}), it is not necessary to send a warning to the driver; if the driver is not aware of the situation and the departure may exceed a certain threshold (i.e., the dashed blue line in Fig. \ref{fig.1}), a warning should be sent to the driver to avoid a crash. Saito \textit{et al}. \cite{saito2016driver} applied the idea in Fig. \ref{fig.1} with a dual control scheme in driver drowsiness. In their study \cite{saito2016driver}, the steering angle was used to estimate the driver's state and then decide whether an assist control should be implemented. Though this approach can improve safety by identifying driver drowsiness, it can not predict the future trajectory of the vehicle and the driver's future operations.

\subsection{Contributions}
We propose an online model-based prediction algorithm of vehicle lateral trajectory to prevent false warnings by inferring whether the driver's forthcoming behavior can bring the vehicle back to the driving lane or cross the lane boundary and potentially cause a crash. The goal is to employ the personalized driver model (PDM) to infer the upcoming lateral trajectory of a vehicle by using a model-based prediction algorithm. We apply a PDM in our research because the driving styles of lane departure and lane keeping behaviors for drivers differ greatly, and the PDM can discern the characteristics of individual's driving style. The proposed PDM describes the driver's lane-keeping behavior and lane-departure behavior by using a joint-probability density distribution of Gaussian mixture model (GMM) between vehicle speed, relative yaw angle, relative yaw rate, lateral displacement, and road curvature. Then, a hidden Markov model (HMM) is used to estimate the vehicle's lateral displacement based on the trained GMM. An online model-based prediction algorithm is developed to predict the vehicle's lateral displacement. We can then predict the future trajectory of the vehicle and judge whether the driver will approach the lane boundary and then bring the vehicle back, keeping the vehicle in the driving lane. Our main contributions are  (1) the personalized driver model (2) and the model-based prediction algorithm.

\subsection{Paper Organization}

The contents of this paper are organized as follows. Section II describes the key concept of the lane-departure predictions and warnings. Section III presents the structure of the proposed approach and model-based prediction algorithm. Section IV describes the experiments and data collection. Section V and VI include a further analysis and a discussion of the results. Section VII gives the conclusions.

\section{Lane Departure Prediction}
Lane departure prediction (LDP) aims to estimate whether a vehicle will depart from the lane, thus allowing a longer time for a driver to take effective action to avoid a crash. The LDP algorithms in the literature can be roughly classified into three groups:  TLC-based prediction\cite{angkititrakul2011use,mammar2006time,cario2009predictive}, vehicle-variable-based vehicle-position estimation\cite{leng2010vision,gaikwad2015lane,dahmani2013road,dahmani2015vehicle}, and detection of the lane boundary using real-time road images\cite{chang2008onboard,cualain2012multiple,lee2002machine}. Some of the methods in the first and second categories share a common feature, namely, they all use real-time images. The vision- or vehicle-variable-based method can improve TLC-based prediction, but have a hard time predicting  DCB, these methods are difficult to deal with it. In this paper, we  focus on predicting vehicle trajectory and deciding whether to send a warning, that is , developing a more effective warning system, instead of lane-detection techniques. 

\subsection{Time to Lane Crossing (TLC)}
\begin{figure}[t]
	\centering
	\includegraphics[scale = 0.9]{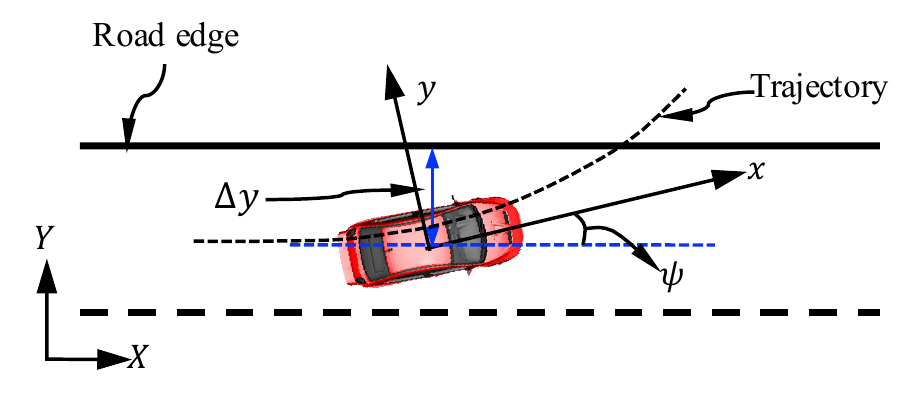}
	\caption{Illustration of a lane departure event on a road.}
	\label{fig.2}
\end{figure}

The TLC-based algorithm has been widely reported in the literature and used on production vehicles. These systems estimate the lane states \cite{dahmani2013road,gaikwad2015lane} (i.e., lane markers, lane width, and lane curvature, etc.) and vehicle states based on vision-based equipment, and then calculate the TLC online using a variety of algorithms\cite{mammar2006time}. When the TLC reaches a threshold, the LDW sends alerts the driver.  The computation methods of TLC differ with regard to various road geometries and vehicle types.  The most common method for calculating the TLC is to predict the road boundary and the vehicle trajectory and then calculate the time when they intersect. When road curvature is small, the TLC can be computed as the ratio of lateral distance to lateral velocity or the ratio of the distance to the line crossing \cite{mammar2006time}:

\begin{equation}\label{eq2.1}
t_{TLC} = \frac{\Delta y - (D/2-l_{f}\cdot \tan\psi)}{v\sin \psi}
\end{equation}
where $ \Delta y $ is the lateral distance from the vehicle's center of gravity (CoG) to the line that would be crossed, $ \psi $ is the relative yaw angle between the longitudinal coordinate of the vehicle and the road direction as shown in Fig. \ref{fig.2}, $ v $ is the vehicle speed, $ D $ is the width of vehicle, and $ l_{f} $ is the distance between the front axle and the CoG of the vehicle. 

\subsection{Excessive False Warning}
\begin{figure}[t]
	\centering
	\begin{subfigure}[t]{0.5\textwidth}
		\centering
		\includegraphics*[scale = 1]{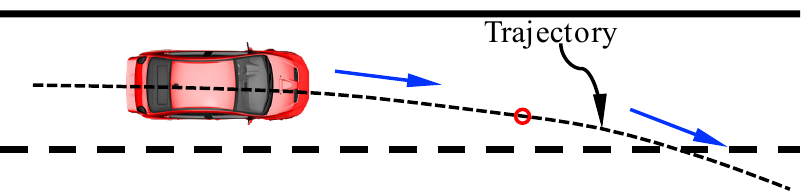}
		\caption{Unintentional lane-departure behavior.}
	\end{subfigure}
	~
	\begin{subfigure}[t]{0.5\textwidth}
		\centering
		\includegraphics*[scale = 1]{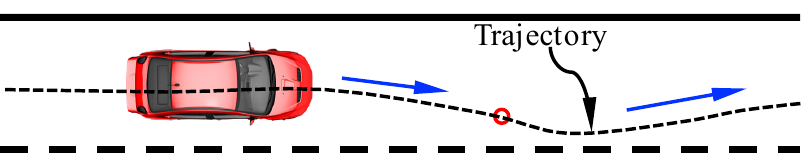}
		\caption{Intentional driver correction behavior.}
	\end{subfigure}
	\caption{Examples of an unintentional lane-crossing event and intentional driver correction behavior. The red circle represents the TLC-based warning methods.}
	\label{fig.3}
\end{figure}

Many studies have observed that the TLC-based methods tend to have a higher FAR when the ego vehicle drives close to the lane boundary\cite{albousefi2016two,angkititrakul2011use,mammar2006time}. This is primarily due to using an oversimplified model to reduce computational complexity, viz. neglecting drivers' steering characteristics and vehicle dynamics. Another reason for the high FAR is that some drivers will control the vehicle back to the center of the lane without the help of a warning with DCB. The problem, however, is that most LDW systems can not predict the forthcoming driver  behaviors or vehicle trajectories. Fig. \ref{fig.3} shows two cases consisting of unintentional LDB and intentional DCB. In the case shown in Fig. \ref{fig.3}(b), an LDW is not desired because the driver can guide the vehicle back after being close to the lane boundary. This kind of false warning is difficult to reduce by improving the accuracy of sensor data (e.g., road curvature) or a TLC-based calculation method (e.g., considering the attributes of vehicle dynamics). Therefore, to reduce this kind of false warning, we need to estimate the driver's forthcoming behavior by asking, ``\textit{Will the driver bring the vehicle back to the center of the driving lane within a short span of time?}'' or ``\textit{Does the driver correctly understand the driving situation?}''. 

\section{Proposed Method}
\begin{figure}[t]
	\centering
	\includegraphics[scale = 0.71]{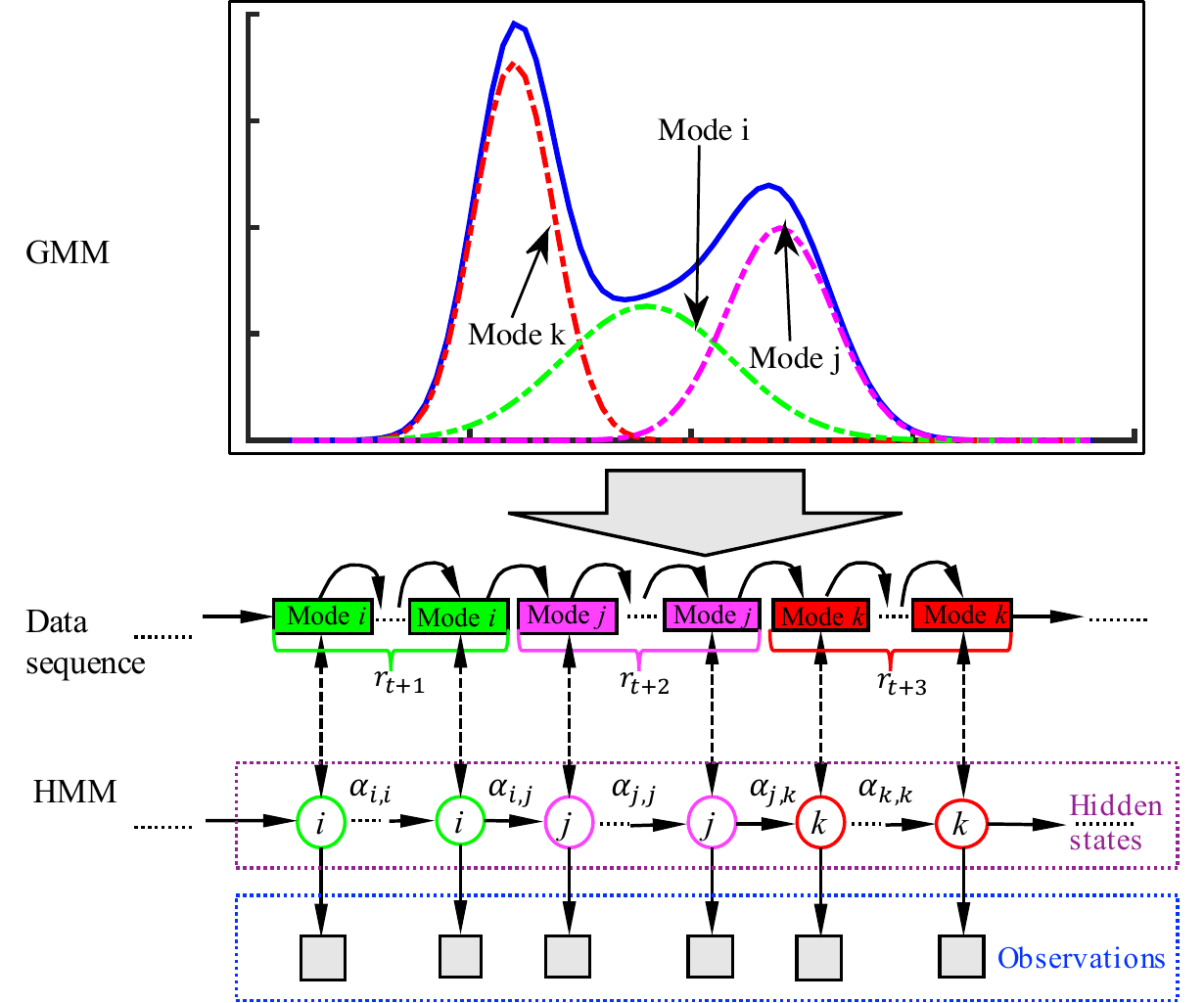}
	\caption{The illustration of the proposed method by combining GMM and HMM, and $ r_{(\ast)} \in \mathbb{N}^{+}$ is the duration for mode $ i,j$, and $ k $.}
	\label{stru_method}
\end{figure}

HMM has been widely used to model driver dynamic behaviors due to its powerful ability to describe the dynamic process and infer unobserved (hidden) states\cite{tang2016modeling}. In the HMM, we apply the component of GMM to representing the hidden modes (Fig. \ref{stru_method}), which makes it easier to determine the Markov chain modes. We use the GMM to model the dependent relationship between variables that could describe driver behaviors because the GMM method has been applied to model driving tasks and has shown its effectiveness \cite{angkititrakul2011use,butakov2015personalized}, thus more off-the-shelf techniques for estimating the model parameters can be directly used. Based on the GMM, we use the observations to infer the hidden states and design a model-based prediction algorithm to infer driver upcoming behaviors.

\subsection{Feature Parameter Selection}
For the LDP, we focus on the driver's lateral-control behaviors. Drivers predict the future trajectory of a vehicle based on their ``internal model'' \cite{ungoren2005adaptive} and the driving situation as well as vehicle states, and then take a lateral-control action to keep the vehicle in a safe and comfortable region. We describe drivers' lane-keeping or lane-departure behaviors using the following variables:

\begin{itemize}
	\item \textit{Vehicle Speed} ($ v $): Speed selections have a great influence on drivers' lateral control \cite{tarko2009modeling} and drivers' risk perception \cite{bella2013driver}. Drivers will usually compensate for steering errors by adjusting the vehicle speed \cite{van1996speed}, keeping the TLC constant. Also, according to Mammar \cite{mammar2006time}, the speed is one key to influencing the TLC. With the same relative yaw angle of $ \psi \neq 0 $, higher speed leads to a smaller TLC.
	\item \textit{Relative Yaw Angle} ($ \psi $): According to May and Baldwin [\ref{eq2.1}], the TLC is also influenced by the relative yaw angle (Fig. \ref{fig.2}). For example, a larger relative yaw angle generates a shorter TLC and has a greater potential for crashes. Therefore, the relative yaw angle is selected as one variable for modeling driver behaviors. 
	\item \textit{Relative Yaw Rate} ($ \dot{\psi} $): The relative yaw rate can indirectly show a driver's intentions. For example, when the vehicle is approaching the road boundary, an inverse relative yaw rate can slow down the approaching speed towards the road boundary and avoid a crash.
	\item \textit{Road Curvature} ($ \rho $): Estimation of road profiles is a big challenge for calculating TLC due to the difficulty of accurately estimating road geometry. In general, to simplify the model and reduce computation cost, road curvature is assumed to be constant or affine varying \cite{mammar2006time} over a short span of time. The means of computing TLC differ with respect to a straight road and a curved road. In this paper, we mainly consider the scenarios where the road has a small curvature $ \rho \leq 10^{-4} $ m$ ^{-1} $.
	\item \textit{Lateral Displacement} ($ \Delta y $): Drivers differ in terms of their preferences for maintaining a vehicle's lateral position with respect to the road boundary, related to many factors such as the driver's ability to perceive risk\cite{bella2013driver} and driver states\cite{ting2008driver}. Drivers can laterally displace the vehicle according to their own physical and psychological states. Therefore, lateral displacement is also selected to describe driver behaviors.
\end{itemize}
In sum, five variables are employed to model drivers' lane-keeping and lane-departure characteristics. We can describe driver behavior using a driving-data sequence, 

\begin{equation*}
\boldsymbol{\xi}_{1:n} = \{  \boldsymbol{\xi}_{1},  \boldsymbol{\xi}_{2}, \cdots,  \boldsymbol{\xi}_{t}, \cdots, \boldsymbol{\xi}_{n}  \},
\end{equation*}
where $ \boldsymbol{\xi}_{t} =  \{ v_{t}, \psi_{t},\rho_{t},\Delta y_{t} , \dot{\psi}_{t} \}$ is the data point at time $ t $ and $ n $ is the number of data point.

\subsection{GMM}
\subsubsection{Structure of GMM} GMM is used to establish the dependent relationships among the five variables in vector $ \boldsymbol{\xi} $. The joint probability density function of $ \boldsymbol{\xi} $ is in a form of the multivariate Gaussian distribution (MGD) function's weighted sum:

\begin{equation}\label{eq3.1}
\begin{split}
p(\boldsymbol{\xi}_{t};\boldsymbol{\theta}) =  & \sum_{k=1}^{K}\omega_{k}\mathcal{N}(\boldsymbol{\xi}_{t};\boldsymbol{\mu}_{k}, \boldsymbol{\Sigma}_{k}) \\
 = & \sum_{k=1}^{K}\omega_{k} \frac{1}{(2\pi)^{d/2}\lvert\boldsymbol{\Sigma}_{k}\rvert^{1/2}} \\
 & \times \exp \left[  -\frac{1}{2}(\boldsymbol{\xi}_{t} - \boldsymbol{\mu}_{k})^{\top}  \boldsymbol{\Sigma}_{k} (\boldsymbol{\xi}_{t} - \boldsymbol{\mu}_{k}) \right] 
\end{split}
\end{equation}
where $ \boldsymbol{\theta} = \{\theta_{k}\}_{k=1}^{K} = \{ \omega_{k},\boldsymbol{\mu}_{k},\boldsymbol{\Sigma}_{k}\}_{k=1}^{K} $ are the parameters of model (\ref{eq3.1}); $ \mathcal{N}(\boldsymbol{\xi}_{t};\boldsymbol{\mu}_{k},\boldsymbol{\Sigma}_{k}) $ is the MGD of $ \boldsymbol{\xi} $, with the mean center $ \boldsymbol{\mu}_{k} \in \mathcal{R}^{d\times 1} $ and covariance matrix $ \boldsymbol{\Sigma}_{k} \in \mathcal{R}^{d \times d} $; $ K $ is the number of GMM components, which can be determined using Bayesian information criterion (BIC) \cite{calinon2010learning}; $ \omega_{k} \in (0,1]$ is the weight of the $ k^{th} $ Gaussian component and $ \sum_{k = 1}^{K} \omega_{k} = 1 $. 

\subsubsection{Parameter Identification} Given a data sequence $ \boldsymbol{\xi}_{1:n} $ and model configuration (\ref{eq3.1}), the model parameter ($ \boldsymbol{\theta} $) can be estimated using the maximum-likelihood (ML) method. The goal of the ML method is to find the parameter $ \boldsymbol{\theta} $ that maximizes the likelihood of the GMM function (\ref{eq3.1}):

\begin{equation}\label{eq3.2}
\mathcal{L}(\boldsymbol{\theta}) = \sum_{t=1}^{n} \log \left(  p(\boldsymbol{\xi}_{t}; \boldsymbol{\theta}) \right) 
\end{equation}
Because of the nonlinearity of (\ref{eq3.2}), it is difficult to directly derive (\ref{eq3.2}) with respect to parameters $ \boldsymbol{\theta} $ and get an optimal solution. Therefore, the iterative version of the Expectation-Maximization (EM) algorithm is employed, which can guarantee a monotonic increase in the likelihood value of model (\ref{eq3.1}) at each step of the iteration, with the objective of searching an optimal parameter $ \boldsymbol{\theta}^{\ast}  $

\begin{equation}\label{eq3.3}
\boldsymbol{\theta}^{\ast} = \arg  \underset{\boldsymbol{\theta}}{\max} \ \mathcal{L}(\boldsymbol{\boldsymbol{\theta}})
\end{equation}

To achieve (\ref{eq3.3}), we denote the EM estimate of $ \boldsymbol{\theta} $ at step $ l $ be $ \widehat{\boldsymbol{\theta}}^{l} $. The iteration from $ \widehat{\boldsymbol{\theta}}^{l} $ to $ \widehat{\boldsymbol{\theta}}^{l+1} $ is achieved by the following \textit{E-Step} and \textit{M-Step}. 

\begin{itemize}
	\item \textit{E-Step}: For each iteration, we compute the posterior probability for each component $ k $ by using the GMM parameter from the previous iteration $ \widehat{\boldsymbol{\theta}}^{l} $:
	
	\begin{equation}\label{eq3.4}
	P_{k}^{l+1}(\boldsymbol{\xi}_{t}) = \frac{\widehat{\omega}_{k}^{l}\cdot\mathcal{N}(\boldsymbol{\xi}_{t};\widehat{\boldsymbol{\mu}}_{k}^{l},\widehat{\boldsymbol{\Sigma}}_{k}^{l})}{\sum_{j=1}^{K}\widehat{\omega}_{j}^{l}\cdot\mathcal{N}(\boldsymbol{\xi}_{t};\widehat{\boldsymbol{\mu}}_{j}^{l},\widehat{\boldsymbol{\Sigma}}_{j}^{l})}
	\end{equation}
	
	\item \textit{M-Step}: Then, update the model parameters by 
	
	\begin{subequations}\label{eq3.5}
		\begin{align}
		\widehat{\omega}_{k}^{l+1} & = \frac{1}{n} \sum_{t=1}^{n} P_{k}^{l+1}(\boldsymbol{\xi}_{t}) \\
		\widehat{\boldsymbol{\mu}}_{k}^{l+1} & = \frac{\sum_{t=1}^{n} (\boldsymbol{\xi}_{t} \cdot P_{k}^{l+1}(\boldsymbol{\xi}_{t}) )}{\sum_{t=1}^{n}  P_{k}^{l+1}(\boldsymbol{\xi}_{t}) } \\
		\widehat{\boldsymbol{\Sigma}}_{k}^{l+1} & = \frac{\sum_{t=1}^{n} \left( P_{k}^{l+1}(\boldsymbol{\xi}_{t}) (\boldsymbol{\xi}_{t} - \widehat{\boldsymbol{\mu}}_{k}^{l+1}) (\boldsymbol{\xi}_{t} - \widehat{\boldsymbol{\mu}}_{k}^{l+1})^{\top}\right) }{\sum_{t=1}^{n}P_{k}^{l+1}(\boldsymbol{\xi}_{t})}
		\end{align}
	\end{subequations}
	
	\item \textit{Update log-likelihood}: At the end of each iteration, we compute and update the log-likelihood $ \mathcal{L}(\widehat{\boldsymbol{\theta}}^{l+1}) $ by
	
	\begin{equation}\label{eq3.6}
	\mathcal{L}(\widehat{\boldsymbol{\theta}}^{l+1}) = \sum_{t=1}^{n} \mathcal{L}(\widehat{\boldsymbol{\theta}}^{l})
	\end{equation}
	
	\item \textit{Convergence Condition}: Repeat the iteration (\ref{eq3.4}) - (\ref{eq3.6}) until the following condition is valid: $ 	\mathcal{L}(\widehat{\boldsymbol{\theta}}^{l+1}) - \mathcal{L}(\widehat{\boldsymbol{\theta}}^{l})  < \varepsilon$, where $ \varepsilon  $ is a very small positive value. In this work, we set $ \varepsilon = 10^{-10} $.
\end{itemize}

\subsection{HMM}
Based on the trained GMM, a connected HMM representation of driver lane-keeping behavior can be built. Each component of GMM is treated as a state in the HMM.  Since the goal is to infer the driver's upcoming behavior based on the driving situation, we define the following variables for the HMM:

\begin{itemize}
	\item \textit{Hidden Modes}: $ m_{t} \in \mathcal{M} =  \{ 1,2, \cdots, K \} $ is the hidden mode at time $ t $, with $ K $ as the number of possible hidden modes;
	\item \textit{Observable States}:  $ \mathcal{O} = \{ \boldsymbol{\zeta}_{t} \}_{t=1}^{n} $ is the set of observable variable states, where $ \boldsymbol{\zeta}_{t} = \{ v_{t}, \psi_{t}, \rho_{t}, \Delta y_{t} \} \in \mathcal{R}^{4\times 1} $  is the observable state at time $ t $;
	\item \textit{Hidden States}: $ \mathcal{H} = \{\dot{\psi}_{t}\}_{t=1}^{n} $ is the set of hidden states, where $ \dot{\psi}_{t} $ is the hidden state that needs to be estimated at time $ t $;
	\item \textit{Transition Matrix}: $ \mathcal{T} = \{ \alpha_{i,j} \} \in \mathcal{R}^{K \times K}$ is the transition matrix, where $ \alpha_{i,j} $ is the transition probabilities from the $ i^{th} $ to $ j^{th} $ hidden modes and $ i,j \in \mathcal{M} $.
\end{itemize}
The transition matrix $ \mathcal{T} $ can be estimated from the training data. See Appendix A.

In the training phase of HMM, the observation $ \boldsymbol{\xi}_{t} = [\boldsymbol{\zeta}_{t},\dot{\psi}_{t} ]$ consists of the observable states and hidden states. The joint distribution between the hidden modes and the observations is presented by 

\begin{equation}
p(m_{0:t}, \boldsymbol{\xi}_{1:t}) = p(m_{0}) \prod_{i=1}^{t} \left[ p(m_{i}|m_{i-1}) \cdot p(\boldsymbol{\xi}_{i}|m_{i}) \right] 
\end{equation}
 Thus, we obtain an MGD, $ p(\boldsymbol{\xi}_{i}|m_{i}) $, with  mode $ m_{i} $ and model parameter $ \theta_{i} $. The model parameter $ \theta_{i} $ can be estimated using the EM algorithm as shown above. In the phase of inferring the hidden states, we will estimate the hidden states (i.e., relative yaw rate) at time $ t $ from the consecutive values of the driving situation using GMM--HMM, i.e., $ \widehat{\dot{\psi}} $ is estimated as the conditional expectation of $ \dot{\psi} $ given the sequence $ \boldsymbol{\zeta}_{1:t} $\cite{lefevre2015driver}:

\begin{equation}\label{eq3.7}
\begin{split}
\widehat{\dot{\psi}}_{t} & = E[\dot{\psi}_{t}|\boldsymbol{\zeta}_{1}, \boldsymbol{\zeta}_{2},\cdots,\boldsymbol{\zeta}_{t} ] \\
 & = \sum_{k=1}^{K}\beta_{k,t}\left[ \boldsymbol{\mu}_{k}^{\dot{\psi}} + \boldsymbol{\Sigma}_{k}^{\dot{\psi}\boldsymbol{\zeta}} (\boldsymbol{\Sigma}_{k}^{\boldsymbol{\zeta}\boldsymbol{\zeta}})^{-1} (\boldsymbol{\zeta}_{t} - \boldsymbol{\mu}_{k}^{\boldsymbol{\zeta}}) \right] 
\end{split}
\end{equation}
where 
\begin{equation*}
\boldsymbol{\mu}_{k} = 
\begin{bmatrix}
\boldsymbol{\mu}_{k}^{\boldsymbol{\zeta}} \\
\boldsymbol{\mu}_{k}^{\dot{\psi}} 
\end{bmatrix}, 
\boldsymbol{\Sigma}_{k} = 
\begin{bmatrix}
\boldsymbol{\Sigma}_{k}^{\boldsymbol{\zeta}\boldsymbol{\zeta}} & \boldsymbol{\Sigma}_{k}^{\boldsymbol{\zeta}\dot{\psi}} \\
\boldsymbol{\Sigma}_{k}^{\dot{\psi}\boldsymbol{\zeta}} & \boldsymbol{\Sigma}_{k}^{\dot{\psi}\dot{\psi}} \\
\end{bmatrix}
\end{equation*}
and $ \beta_{k,t} $ is the mixing weight for mode $ k $ at time $ t $, computed as the probability of being in mode $ k $ and observing the sequence $ \boldsymbol{\zeta}_{1:t} $. The computation of $ \beta_{k,t} $ is given by 

\begin{equation}\label{eq3.8}
\beta_{k,t} = \frac{\left( \sum_{j=1}^{K} \beta_{k,t-1} \cdot \alpha_{j,k}\right) \cdot \mathcal{N}\left(  \boldsymbol{\zeta}_{t}; \boldsymbol{\mu}_{k}^{\boldsymbol{\zeta}}, \boldsymbol{\Sigma}_{k}^{\boldsymbol{\zeta}\boldsymbol{\zeta}} \right) }{\sum_{r=1}^{K} \left[ \left( \sum_{j=1}^{K} \beta_{k,t-1} \cdot \alpha_{j,r}\right) \cdot \mathcal{N}\left(  \boldsymbol{\zeta}_{t}; \boldsymbol{\mu}_{r}^{\boldsymbol{\zeta}}, \boldsymbol{\Sigma}_{r}^{\boldsymbol{\zeta}\boldsymbol{\zeta}} \right)\right] }
\end{equation}
with initialized value  
\begin{equation*}
\beta_{k,1} = \frac{\omega_{k} \cdot\mathcal{N} (\boldsymbol{\zeta}_{1}; \boldsymbol{\mu}^{\boldsymbol{\zeta}}_{k},\boldsymbol{\Sigma}^{\boldsymbol{\zeta}\boldsymbol{\zeta}}_{k})}{ \sum_{j=1}^{K} \left[ \omega_{j} \cdot \mathcal{N} (\boldsymbol{\zeta}_{1}; \boldsymbol{\mu}^{\boldsymbol{\zeta}}_{j},\boldsymbol{\Sigma}^{\boldsymbol{\zeta}\boldsymbol{\zeta}}_{j})\right] }
\end{equation*}

\subsection{Iterative Algorithm for Prediction}
The proposed method must be able to predict the future trajectory based on the historical information $ \boldsymbol{\zeta}_{1:t} $. This prediction is computed by iteratively applying the driver model defined in (\ref{eq3.7}) and  propagating the driving situation. The prediction algorithm is based on a kinematic point mass model for the ego vehicle and some assumptions as follows:

\begin{itemize}
	\item Over a short period of time, vehicle-speed changes are small and can be treated as a constant for the lane-departure behavior, i.e., the vehicle speed is constant during the iteration process of prediction. The emergent brake behaviors, such as collision avoidance, are not considered.
	\item Road curvature is continuous and can be differentiated. It changes slowly during the iteration process of prediction. To reduce calculation complexity, we consider driving scenarios on a road with a small curvature ($ \rho < 10^{-4} $ m$ ^{-1} $). Therefore, road curvature can be treated as a constant during prediction.
%
%
%
\end{itemize}
Based on the above assumptions, the prediction algorithm is shown in \textbf{Algorithm \ref{Algorithm1}}, in which $ \Delta t $ is the discretization time of the driver model and set $ \Delta t = 0.1 $ s. According to the assumptions made above, the vehicle speed and road curvature are constant during the prediction steps  in \textbf{Algorithm \ref{Algorithm1}}. See (\ref{eq3.12}).

\begin{algorithm}[htbp]
	\begin{algorithmic}[1]
		\State Based on the observation $ \boldsymbol{\xi}_{t} $ at time $ t $, to predict the relative lateral position $ \Delta \widehat{y}_{t+(q+1)\Delta t} $ at time $ t+(q+1)\Delta t$, $ q \in \mathbb{N} $;
		\State Set $ i = 0 $ and define $ \Delta\psi_{t} = \dot{\psi}_{t} $;
		\While{$ i \leq q$ \& $ i \in \mathbb{N} $}
		\State Compute the relative yaw angle $ \psi_{t+(i+1)\Delta t} $ at time $ t+(i+1)\Delta t$ by 
		\begin{equation}\label{eq3.9}
		\psi_{t+(i+1)\Delta t} = \psi_{t+i\cdot\Delta t} + \Delta\psi_{t+i\cdot\Delta t}\cdot \Delta t 
		\end{equation}
		\State Compute the lateral displacement $ \Delta y _{t+(i+1)\Delta t} $ at time $ t+(i+1)\Delta t $ by
		\begin{equation}\label{eq3.10}
		\Delta y_{t+(i+1)\Delta t} = \Delta y_{t+i\cdot\Delta t} + v_{t+i\cdot\Delta t}\cdot\sin \psi_{t+i\cdot\Delta t} \cdot\Delta t
		\end{equation}
		\State Estimate the relative yaw rate $ \dot{\psi}_{t+(i+1)\Delta t} $ at time $ t+(i +1)\Delta t$ according to (\ref{eq3.7}) by
		\begin{equation}\label{eq3.11}
		\begin{split}
		\boldsymbol{\zeta}_{t+(i+1)\Delta t} & \xrightarrow{(\ref{eq3.7})}  \widehat{\dot{\psi}}_{t+(i+1)\Delta t} \\
		\dot{\psi}_{t+(i+1)\Delta t} & \approx \widehat{\dot{\psi}}_{t+(i+1)\Delta t} \\
		\Delta\psi_{t+(i+1)\Delta t} & = \dot{\psi}_{t+(i+1)\Delta t}
		\end{split}
		\end{equation}
		\State Update speed and curvature value at time $ t+(i+1)\Delta t $
		\begin{equation}\label{eq3.12}
		\begin{split}
		v_{t+(i+1)\Delta t} & = v_{t+i\cdot\Delta t} \\
		\rho_{t+(i+1) \Delta t} & = \rho_{t+i\cdot\Delta t} 
		\end{split}
		\end{equation}
		\State Assign  $  \Delta \widehat{y}_{t+(i+1)\Delta t}  \Leftarrow  \Delta y_{t+(i+1)\Delta t}  $ 
		\State Update $ i = i+1 $;
		\EndWhile
		\State Return the estimated relative lateral position $ \Delta \widehat{y}_{t+(q+1)\Delta t} $ at time $ t+(q+1)\Delta t$.
	\end{algorithmic}
	\caption{Model-based Prediction algorithm for $ q $-step prediction.}
	\label{Algorithm1}
\end{algorithm}

\section{Experiments and Data Collection}
\subsection{Data Collection}
\begin{figure}[t]
	\centering
	\includegraphics[width = 0.48 \textwidth]{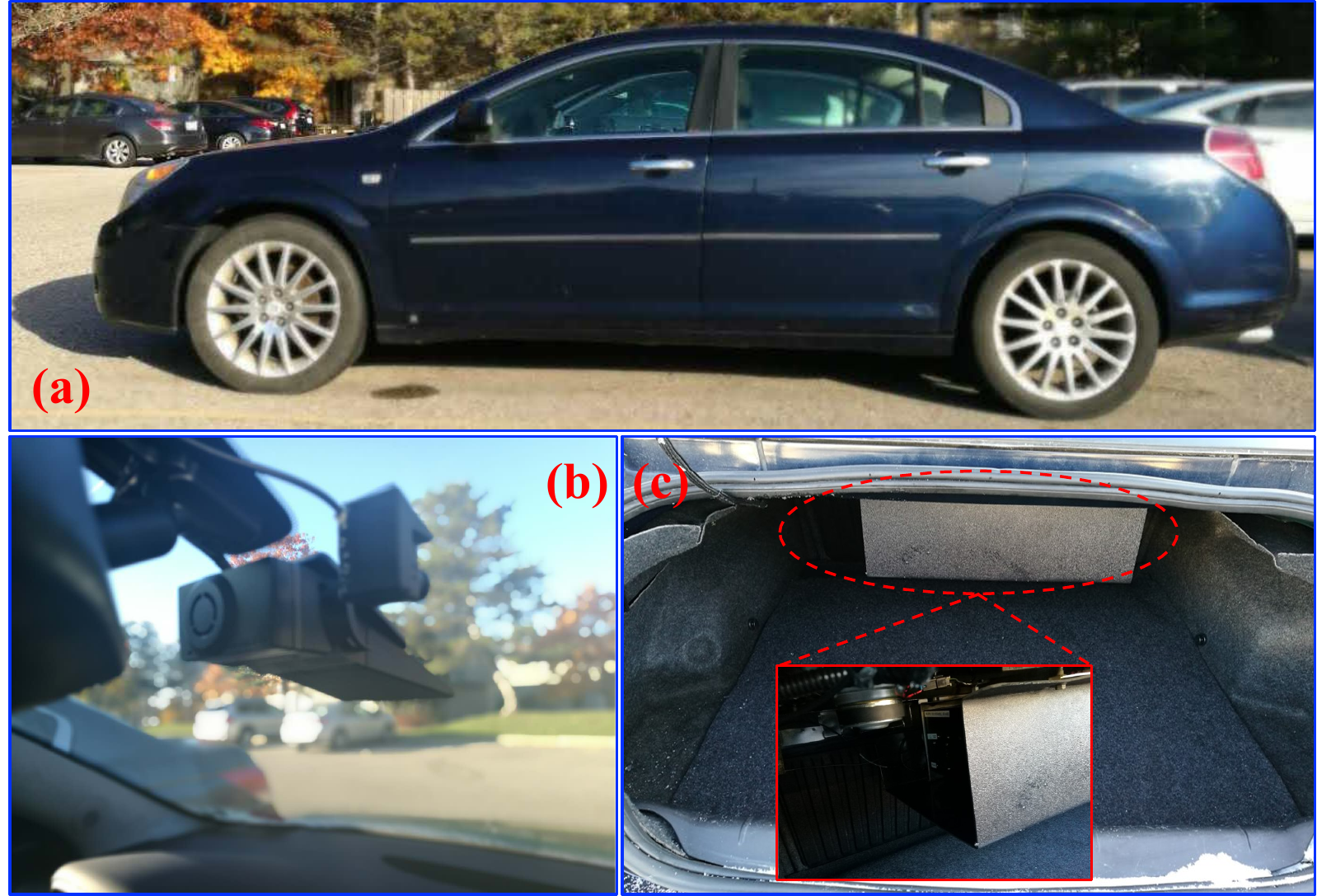}
	\caption{Examples of the experiment vehicle with data-collection equipment. (a) Example of an experiment vehicle; (b) Mobileye; and (c) DAS.}
	\label{exps}
\end{figure}

All the data come from naturalistic driving and are collected from the Safety Pilot Model Deployment program\cite{bezzina2014safety}. Volunteers were recruited from Ann Arbor city with their own vehicles mounted with Mobileye, DSRC, and a data acquisition system (DAS) (Fig. \ref{exps}). The vehicle-based variables such as vehicle speed and the positions of the gas/brake pedal were obtained from the CAN-Bus and the road-based variables are collected from the Mobileye systems. 


\begin{table}[t]
	\centering
	\caption{Statistical Results of the Extracted Driving Data for 10 Drivers}
	\begin{tabular}{c|c|c|c}
		\hline
		\hline
		Driver & \# of events & Total time [min] & Average time [s]\\
		\hline
		1 & 696 & 247.80 & 21.37\\
		2 & 481 & 181.69 & 22.66\\
		3 & 2177 & 759.52 & 20.93\\
		4 & 1861 &  705.76 & 22.75\\
		5 & 3529 & 1244.43 & 21.16\\
		6 & 6048 & 2238.01 & 22.20\\
		7 & 331 & 118.70 & 21.52\\
		8 & 4806 & 1698.36 & 21.20\\
		9 & 4285 & 1504.66 & 21.07\\
		10& 4411 & 1617.04 & 22.00\\
		\hline
		Average & - & - & 21.60 \\
		\hline\hline
	\end{tabular}
	\label{Table1}
\end{table}

Data recorded from ten drivers were used in the experiments. Drivers had an opportunity to become accustomed to the equipped vehicles. They performed casual daily trips for several months without any restrictions on or requirements for their trips, the duration of the trips, or their driving style. While the vehicle was running, the onboard PC recorded driving data with a frequency of 10 Hz. The data process and recording equipment were hidden from the drivers, thus avoiding the influence of recorded data on driver behavior.

\subsection{Data Preprocessing}
The data sets for training a PDM were extracted from the entire set of data. The training data consisted of many \textit{events} that included the DCB and/or LDB. The following rules were considered to determine the beginning points and endpoints for each event of interest:
\begin{itemize}
	\item Detect the \textit{case} data points $ \boldsymbol{\xi} $ in which the vehicle lateral position are close to the road boundary with $ \Delta y \leq 0.5  $ m  or crossing the road boundary \cite{saito2016driver}.
	\item The data points that were behind the  \textit{case} data points with 15 s and before the \textit{case} data point with 15 s were extracted.
	\item The data points with road curvature $ \rho > 10^{-4} $ m$ ^{-1} $ were deleted.
	\item Only the data with road width of about 3.7 m was considered.
	\item We deleted the event with turning lights on, avoiding the lane change behaviors. Also, we checked the event where the lateral displacement changes from one lane to the center of the adjacent lane was deleted. This can avoid some lane change behaviors without turning lights on. 
%
%
	\item The event with a duration of less than 15 s was deleted.
\end{itemize}
The statistical results of the driving data for 10 drivers are listed in Table \ref{Table1}. The number of events ranges from 330 to 6000 and the driving-time duration for each event averages about 22 s.

\begin{figure}[t]
	\centering
	\includegraphics[scale = 0.66]{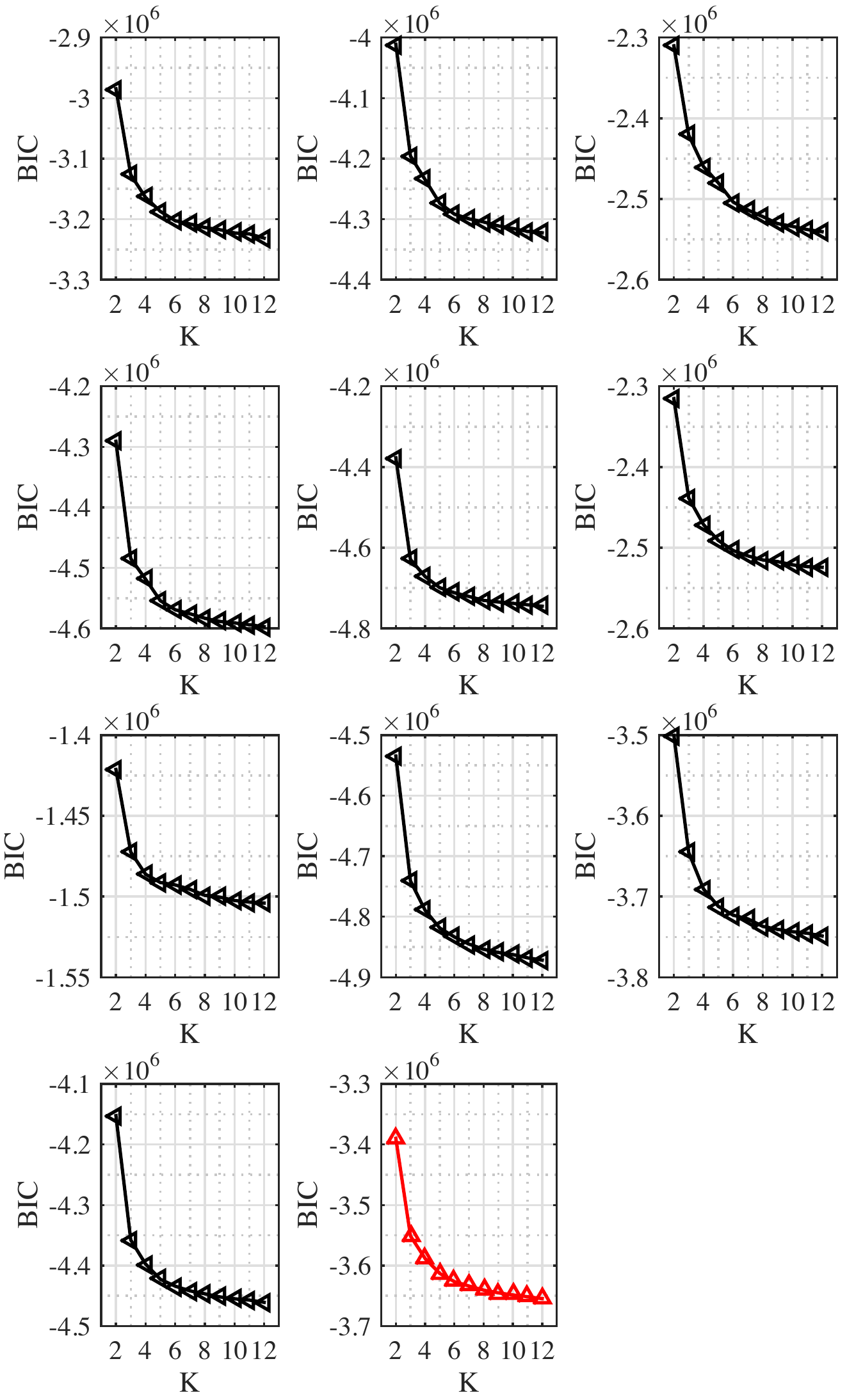}
	\caption{BICs with respect to different numbers of GMM component for 10 drivers (black line with triangle) and the average value of BIC (red line with triangle).}
	\label{BIC}
\end{figure}

\subsection{Training Model}

  \begin{figure*}[t]
  	\centering
  	\includegraphics[scale = 0.66]{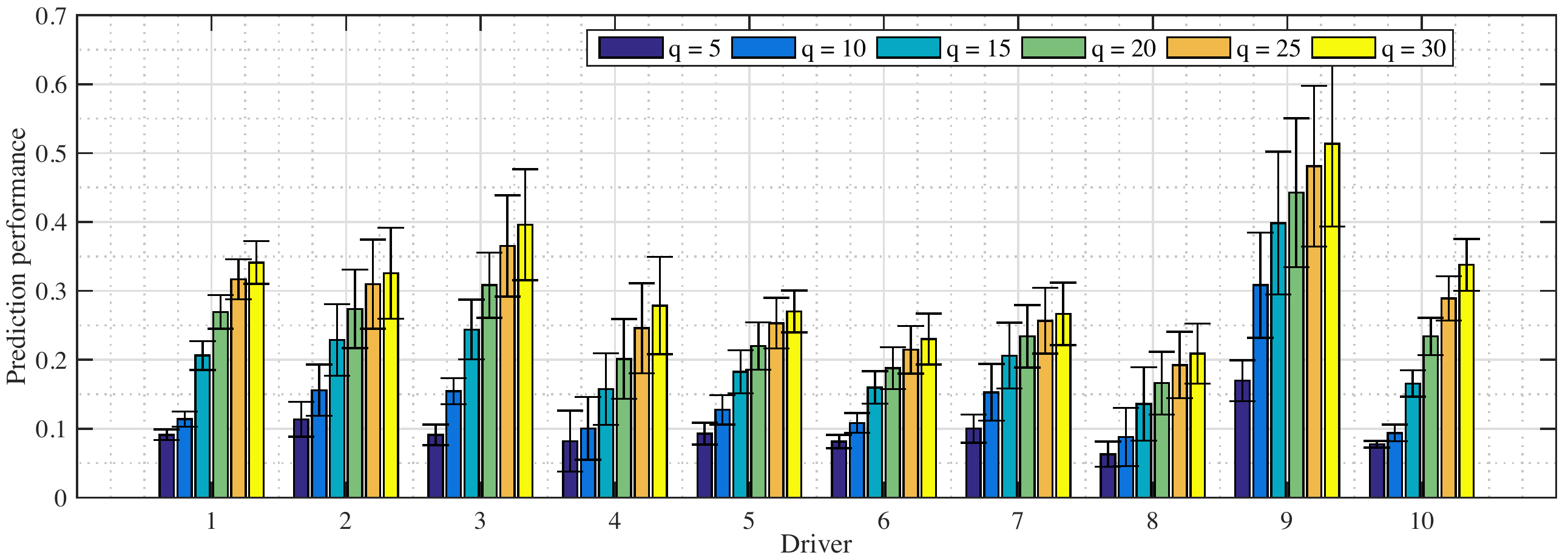}
  	\caption{The errors and stand deviations for 10 drivers with different prediction $ q $-steps. }
  	\label{fig5.1}
  \end{figure*}

 \begin{figure}[t]
 	\centering
 	\includegraphics[scale = 0.62]{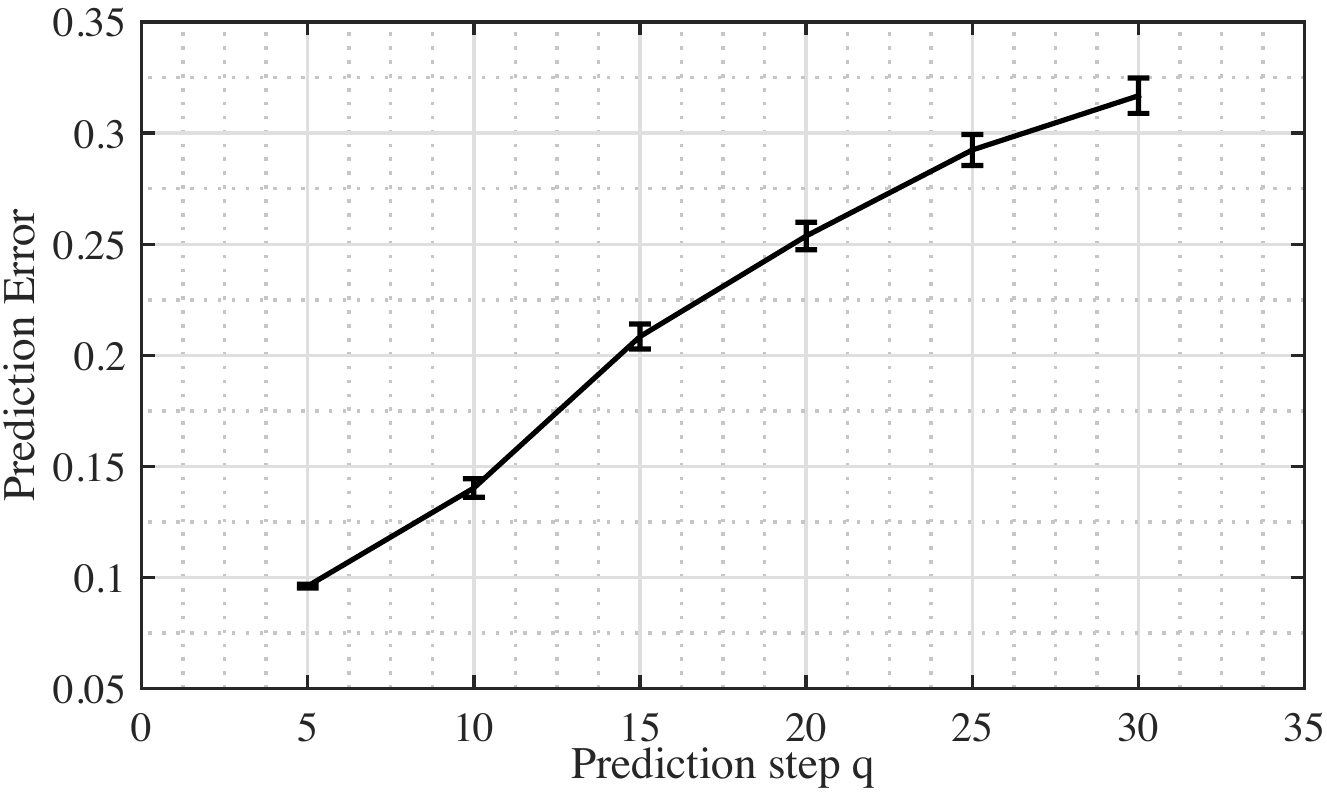}
 	\caption{Prediction errors on average of 10 drivers with respect to different steps $ q $.}
 	\label{fig5.0}
 \end{figure}

A cross-validation (CV) method was selected to train the PDM. Driving data for each driver was evenly divided into ten folds. Nine folds  were used to train the driver model, and the remaining one was used to assess the performance. The CV method guarantees that the training data and test data are not joint. 

We used BIC to determine the number of GMM components by finding the ``elbow point''. Fig. \ref{BIC} presents the experiment results of BIC for different numbers of GMM component in 10 drivers. As the BIC is convergent at about $ K=10 $, we select $ K=10 $.

\subsection{TLC-Based LDP with PDM}
The approach proposed in this paper is named as a TLC-PDM method because it is based on a PDM.  To reduce the FAR, we designed a personalized warning strategy based on the predicted vehicle trajectory. Assuming that we can estimate the relative lateral position of a vehicle at the upcoming time $ t+q \Delta t$ by the \textbf{Algorithm \ref{Algorithm1}}, we can also know whether drivers will bring the vehicle back from road boundaries in the upcoming behaviors, keeping vehicles in the driving lane. Only when the TLC reaches a predefined criterion, is a warning sent, i.e.,

\begin{subequations}\label{eq4.1}
	\begin{align}
	t_{TLC} & < \tau \\
	\min \{\Delta \widehat{y}_{t:t+q\Delta t}\} &  < \gamma_{1} \\ 
	\Delta \widehat{y}_{t+q\Delta t} & < \gamma_{2}
	\end{align}
\end{subequations}
 If the driving conditions (\ref{eq4.1}a) -- (\ref{eq4.1}c) are valid, a warning is sent to drivers for the TLC-PDM method. Equation (\ref{eq4.1}a) checks the TLC condition, (\ref{eq4.1}b) detects the upcoming vehicle trajectory in time [$ t:t+q\Delta  t $] and (\ref{eq4.1}c) judges whether drivers will bring the vehicle back to the center of the lane at future time $ t+q\Delta t $. Thus, only when the TLC is valid and the minimum relative lateral position during future time [$ t:t+q\Delta  t $] is less than a value $ \gamma_{1} $ as well as the vehicle trajectory at future time [$ t+q\Delta t $] is less than $ \gamma_{2} $, is a warning sent.  For example, for $ \gamma_{1} = -0.1 $, $ \gamma_{2} = 0.1 $ and $ q = 30 $, even though the condition (\ref{eq4.1}a) was valid,  if the upcoming trajectories  [$\Delta \widehat{y}_{ t:t+3} $] are always larger than $ -0.1 $ m ( i.e., the driver will not cross the road boundary a lot)\textit{} or the future position $ \Delta \widehat{y}_{t+3}$ is larger than 0.1 m (i.e., the driver will bring the vehicle back to the center of driving lane), a warning will not be sent. Conditions (\ref{eq4.1}b) and (\ref{eq4.1}c)  give drivers more time to bring the vehicle back and keep the vehicle in the driving lane, instead of giving them a warning and being a nuisance. Parameter $ \tau $ is a threshold value of TLC. In \cite{saito2016driver}, the authors agreed that the time margin should be larger than 0.9 s (i.e., $ \tau > 0.9 $ s), thus allowing enough time for a response and applying a correct reaction for drivers. Thus, in this paper, we set $ \tau = 1.0 $ s.
 
  \begin{figure*}[t]
  	\centering
  	\includegraphics[scale = 0.62]{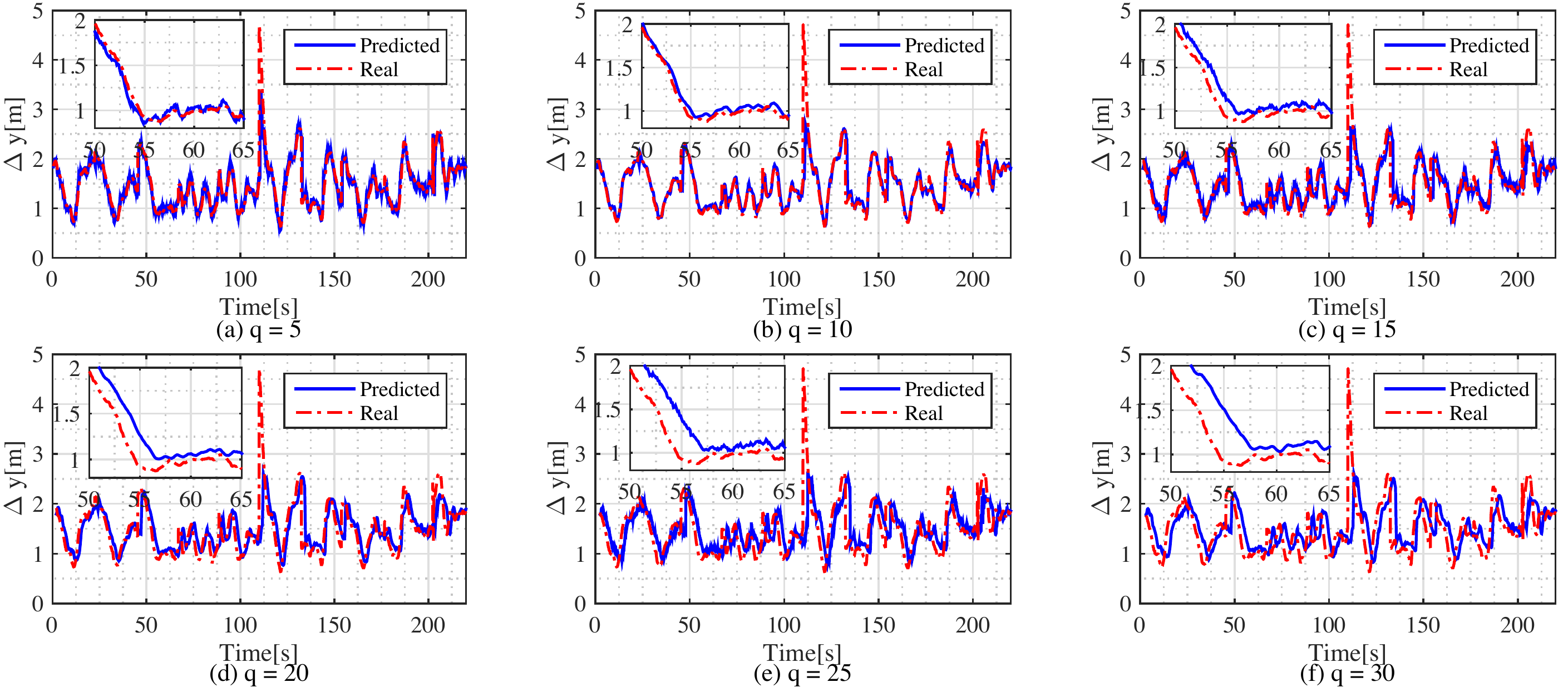}
  	\caption{One example of the prediction results of the relative lateral position $ \Delta y $ for driver \#10 with different prediction steps using the TLC-PDM method.}
  	\label{fig5.2}
  \end{figure*}
  
  \begin{figure}[t]
  	\centering
  	\includegraphics[scale = 0.56]{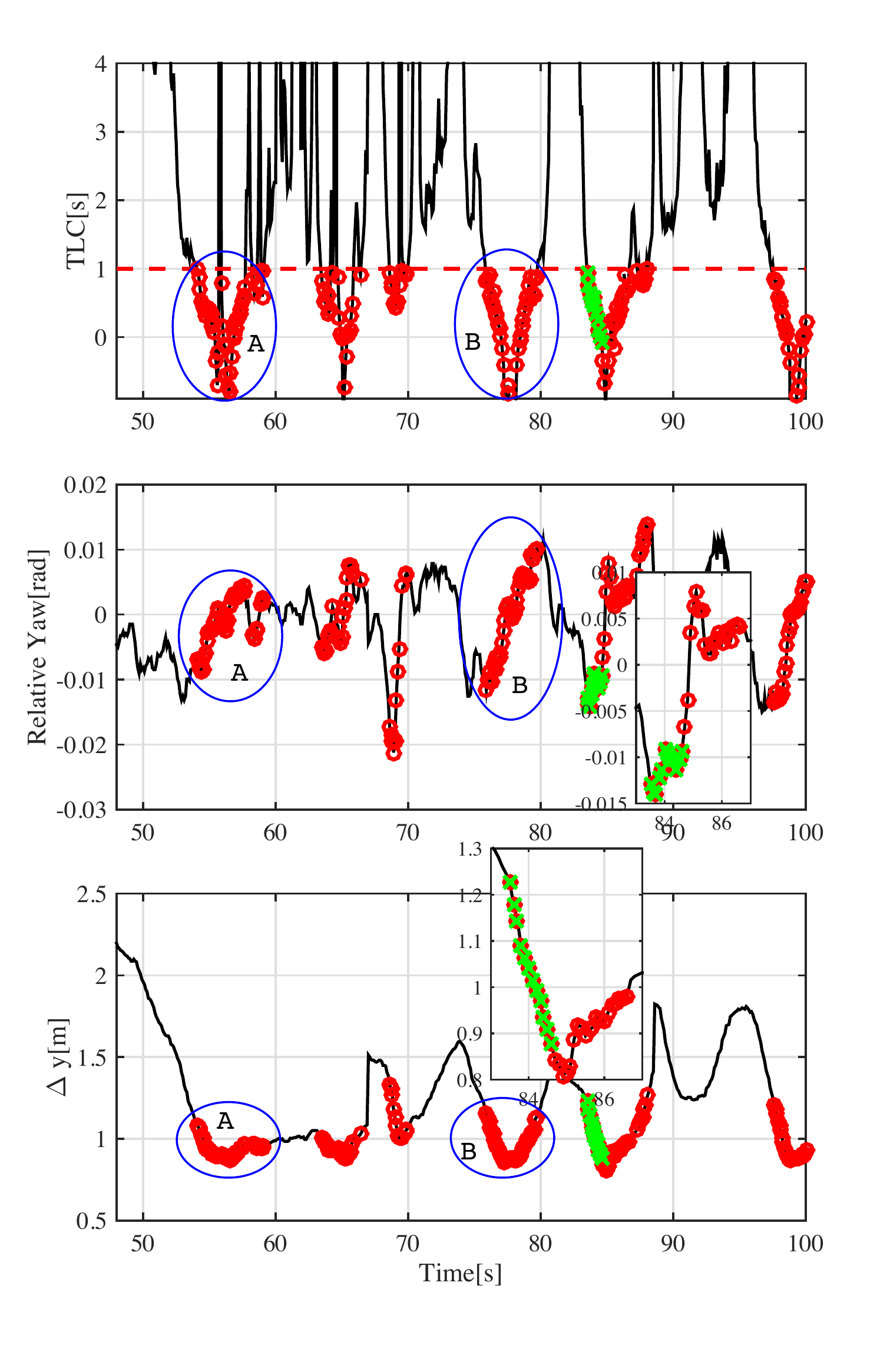}
  	\caption{Example of LDWS using the basic TLC algorithm with $ \tau = 1 $ s and TLC-PDM method with $ \gamma_{1} = -0.05 $, $ \gamma_{2} = 0.1 $, $ q = 10 $ for driver \#10. The red circle represents that (\ref{bTlc}) is valid for the basic TLC algorithm. The green cross represents that the TLC-PDM criteria (\ref{eq4.1}a) -- (\ref{eq4.1}c)  are valid. }
  	\label{fig5.3}
  \end{figure}
 
\subsection{Comparison with Other Approaches}
We compare the TLC-PDM approach with two recent methods that have been proven to demonstrate good performance in LDW applications. One is the basic TLC-based method, which considers no human factors or features of vehicle dynamics. The second one is the directional sequence of the piecewise lateral slopes (DSPLS) method \cite{angkititrakul2011use}, called the TLC-DSPLS method.

\subsubsection{Basic TLC} An alarm signal is given to drivers when the following condition (\ref{bTlc}) is valid. 

\begin{equation}\label{bTlc}
	t_{TLC} < \tau 
\end{equation}
This method is very simple but relatively efficient in a simple driving scenario.
\subsubsection{TLC-DSPLS} TLC-DSPLS method \cite{angkititrakul2011use} is also based on  a probabilistic driver model, in which the trajectory of driving signals is described by the DSPLS which then infers drivers' upcoming episodes of vehicle trajectory. The Bayes rule is used to compute the probabilities of $ p(LCB|state) $ and $ p(DCB|state) $. The decision of the occurrence of LCB is given by

\begin{equation}\label{eq4.2}
\frac{p(DCB|state)}{p(LCB|state)} < \gamma_{0}
\end{equation}
 where $ \gamma_{0} $ is a predefined threshold.  In this paper, we just reproduce and re-examine this method as was done in \cite{angkititrakul2011use} to show the benefits of our proposed method. We will not further discuss TLC-DSPLS. More details can be found in \cite{angkititrakul2011use}. Thus, warning if both (\ref{bTlc}) and (\ref{eq4.2}) are valid.
%
%
%

\section{Results and Analysis}
We discuss and analyze the experimental results in terms of the prediction performance of lateral positions and the FAR of the LDW system for DCB. 

\subsection{Prediction of Lateral Positions}
Being able to predict the performance of the vehicle's upcoming trajectories is crucial for designing a personalized LDW system and reducing the FAR. The performance criterion is computed using the errors of predicted and experimental vehicle trajectory:

\begin{equation}\label{eq4.3}
e_{\Delta y_{t}} = \frac{1}{q}\sum_{t}^{t+q}   \left|  \Delta \widehat{y}_{t:t+q\Delta t} - \Delta y_{t:t+q\Delta t} \right| 
\end{equation}

Fig. \ref{fig5.1} presents the prediction results of the vehicle trajectory. We know that the proposed method can predict the vehicle lateral trajectory precisely across different prediction steps. For $ q = 5 $ (i.e., prediction time $ q\Delta t = 0.5 $ s), the prediction error ranges from $ 0.063 $ (driver \#8) to $ 0.1696 $ (driver \#9). For $ q = 30 $ (i.e., prediction time $ q\Delta t = 3.0 $ s), the prediction error ranges from $ 0.2090 $ (driver \#8) to $ 0.5138 $ (driver \#9).  For the same driver, we note that the prediction performance decreases as the prediction step increases. Fig. \ref{fig5.0} shows the average value of prediction errors with respect to prediction step $ q $ for 10 drivers.  Fig. \ref{fig5.2} gives an example of predicted lateral position with respect to different prediction steps. We note that a larger prediction time   (i.e., $ q = 30 $) leads to a larger prediction error. This error could be influenced by the assumption that during iteration of each step in the prediction horizon in \textbf{Algorithm \ref{Algorithm1}}, the road curvature and vehicle speed are treated as constant.
%
%

\begin{figure}[t]
	\centering
	\begin{subfigure}[t]{0.24\textwidth}
		\centering
		\includegraphics[scale = 0.52]{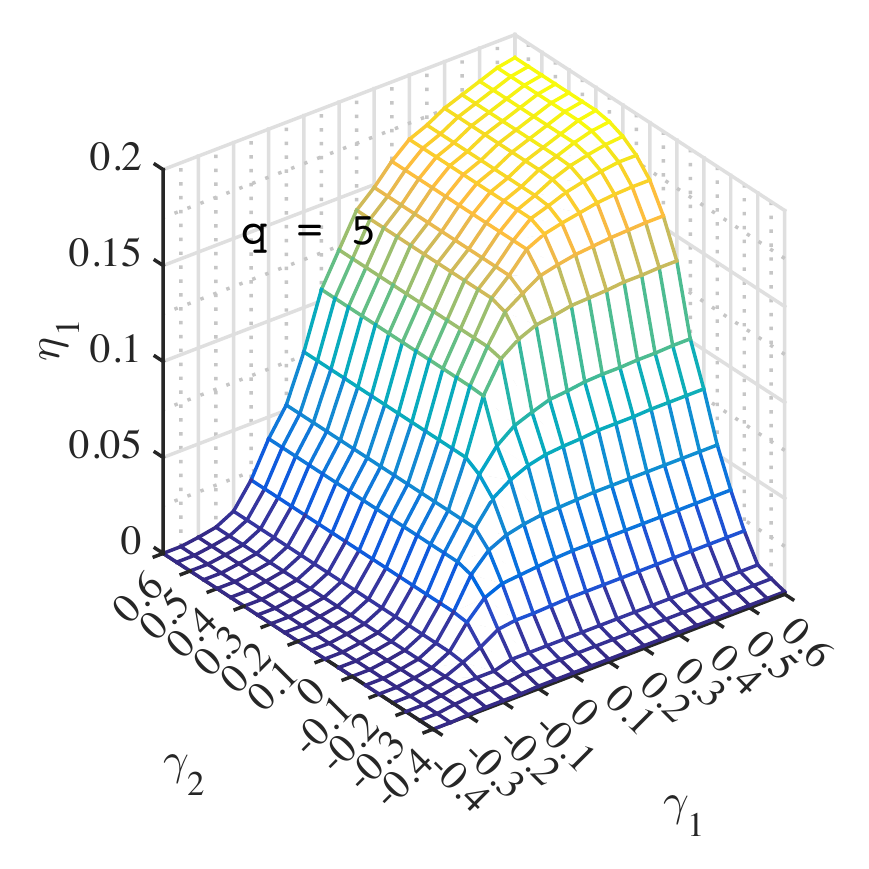}
		\caption{$ q = 5 $}
	\end{subfigure}
	\begin{subfigure}[t]{0.24\textwidth}
		\centering
		\includegraphics[scale = 0.52]{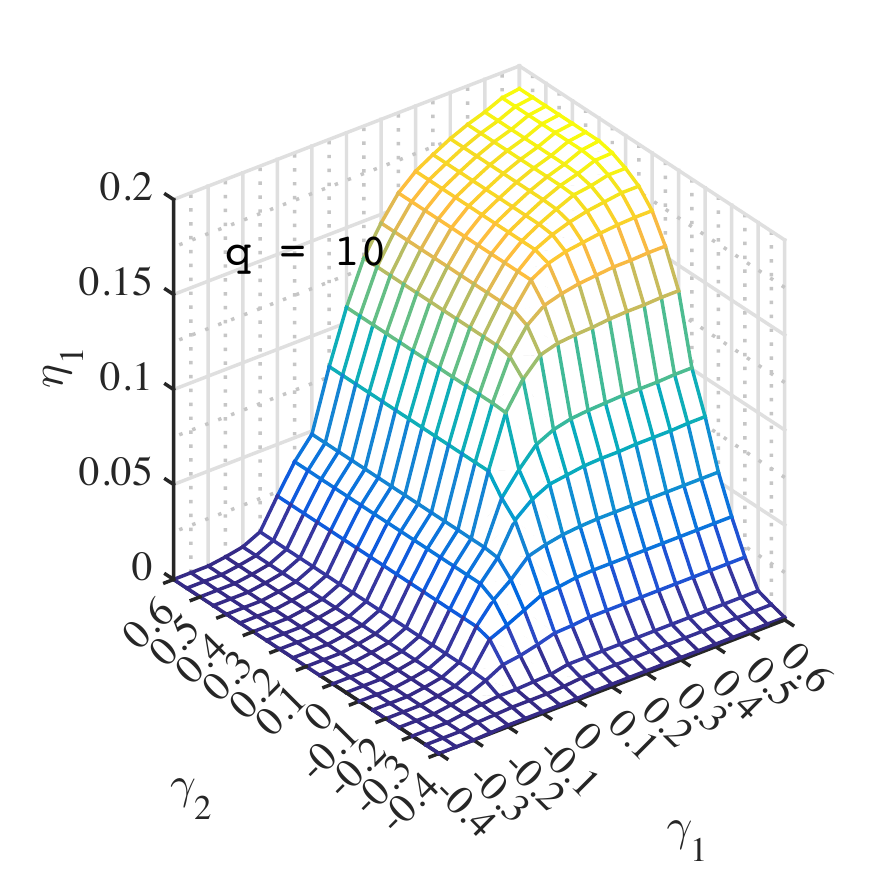}
		\caption{$ q = 10 $}
	\end{subfigure}
	\begin{subfigure}[t]{0.24\textwidth}
		\centering
		\includegraphics[scale = 0.52]{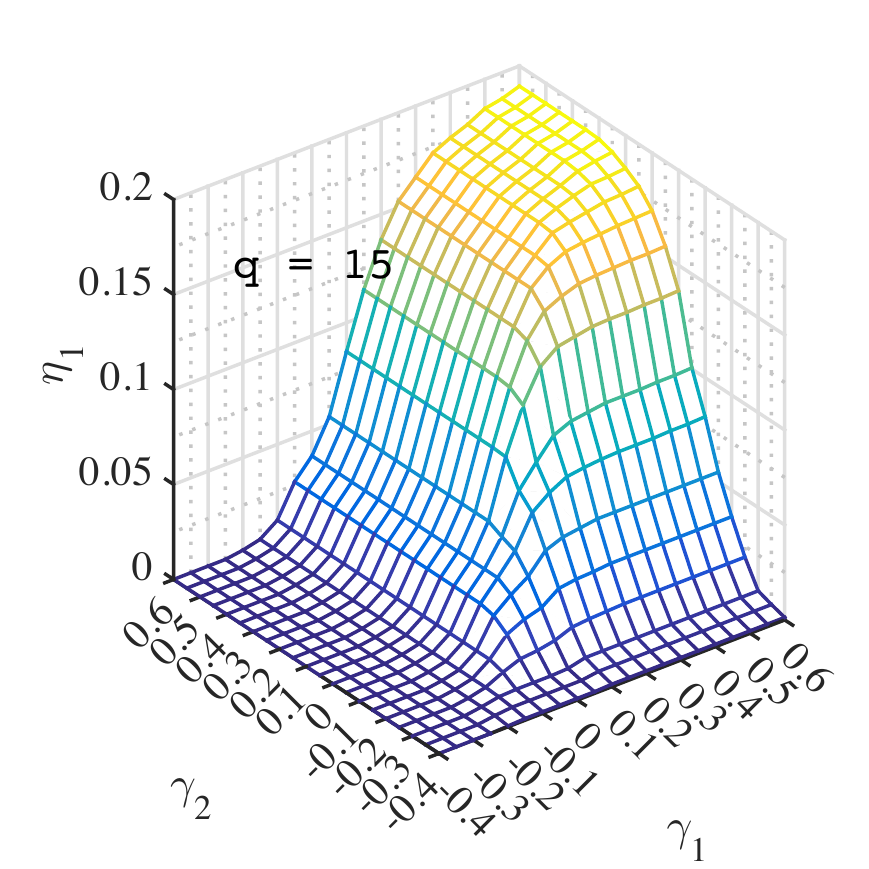}
		\caption{$ q = 15 $}
	\end{subfigure}
	\begin{subfigure}[t]{0.24\textwidth}
		\centering
		\includegraphics[scale = 0.52]{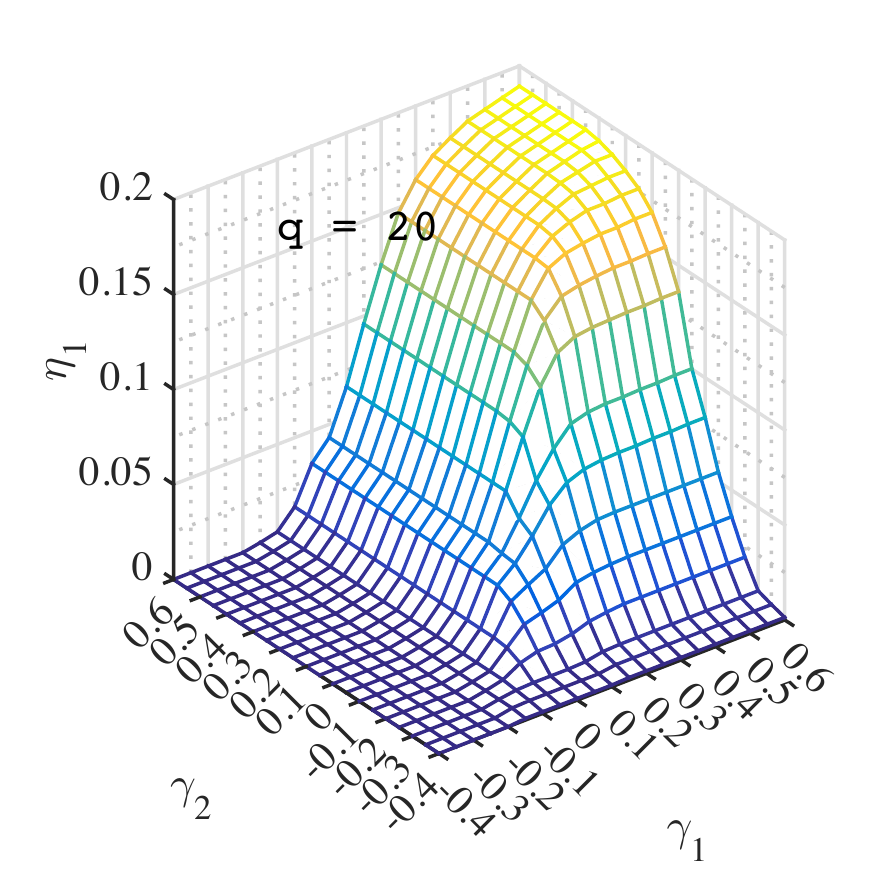}
		\caption{$ q = 20 $}
	\end{subfigure}
	\begin{subfigure}[t]{0.24\textwidth}
		\centering
		\includegraphics[scale = 0.52]{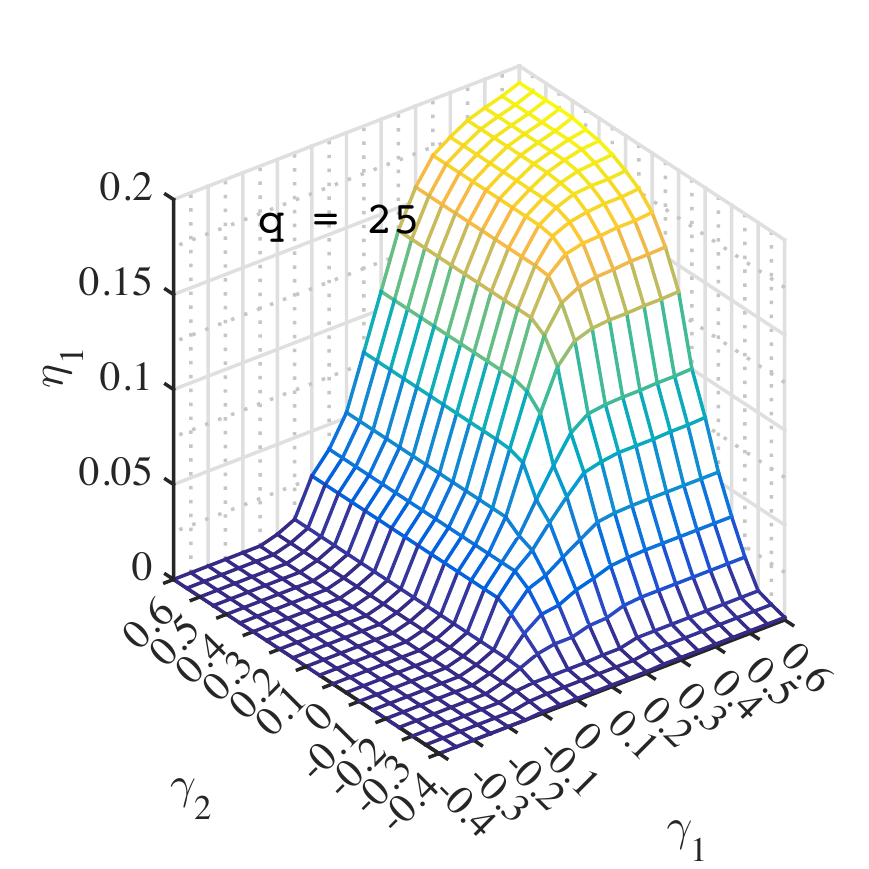}
		\caption{$ q = 25 $}
	\end{subfigure}
	\begin{subfigure}[t]{0.24\textwidth}		
    \centering
    \includegraphics[scale = 0.53]{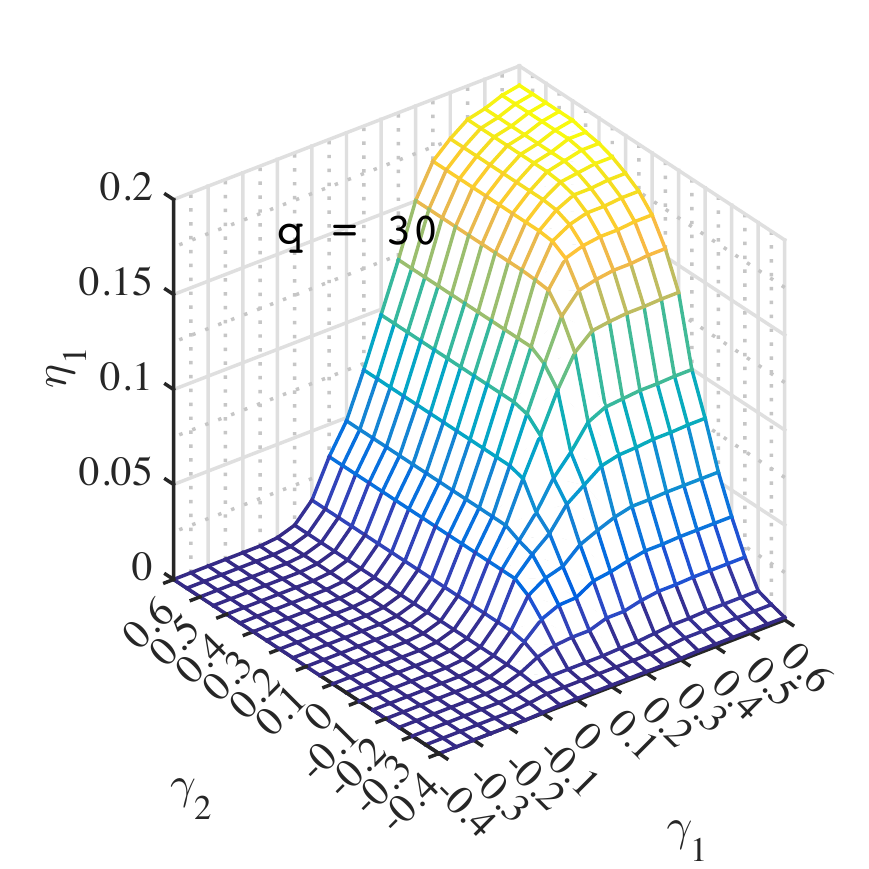}
	\caption{$ q = 30 $}
	\end{subfigure}
 	\caption{Example of the influences of parameters $ \gamma_{1} $ and $ \gamma_{2} $ on the warning over different steps $ q $ by using the TLC-PDM method for driver \#10.}
\label{fig5.4}
 \end{figure}
 
 \subsection{Performances for LDW system}
Fig. \ref{fig5.3} presents an example comparing different LDW algorithms using the basic TLC method and the TLC-PDM method for driver \#10. We note that the basic TLC method is unable to predict or infer whether the driver will bring the vehicle back to the center of the driving lane. Thus, the LDW system will send a warning once condition (\ref{eq4.1}a) is satisfied, even if the driver intends to bring the vehicle back before or after the warning. Furthermore,  we can note that a larger critical value ($ \tau $) of TLC leads to a higher FAR, which tends to annoy drivers.

For the TLC-PDM method, we note that in region \texttt{A} in Fig. \ref{fig5.3}, the driver does not receive a warning from the LDW system, even though the vehicle is laterally crossing the boundary a little (i.e., less than 0.05 m) because the TLC-PDM method can predict that the driver can steer the vehicle back to the center of the driving lane by him/herself in a short time. Region \texttt{B} is the same case as Region \texttt{A}. Green cross symbols in Fig. \ref{fig5.3} indicate that the TLC-PDM criteria (i.e., (\ref{eq4.1}a) - (\ref{eq4.1}c)) are valid and a warning is triggered, because \textbf{Algorithm 1} can predict the vehicle lateral trajectory in the upcoming 1s and estimate that (1) the vehicle will obviously  cross the lane boundary (i.e., $ \min \{\Delta \widehat{y}_{t:t+q\Delta t}\} < \gamma_{1}$ ), and (2) the driver will not be able to bring the vehicle back to the current lane in a short time (i.e., $ \Delta \widehat{y}_{t+q\Delta t} < \gamma_{2} $ ). Therefore, the TLC-PDM method gives the driver a warning, avoiding a crash and reducing the FAR. The TLC-PDM method considers drivers' upcoming behaviors and provides a more acceptable warning decision (i.e., criteria  (\ref{eq4.1}a) -- (\ref{eq4.1}c)). Furthermore, for the TLC-PDM method, parameters $ q $, $ \gamma_{1} $ and $ \gamma_{2} $ in (\ref{eq4.1}a) -- (\ref{eq4.1}c) play a major role in determining  when the LDW system will send a warning. 

\subsubsection{Influences of parameters $ q, \gamma_{1}, \gamma_{2} $ } From (\ref{eq4.1}a) --   (\ref{eq4.1}c), we know that the TLC-PDM warning strategy is developed from the basic TLC warning condition. Different values of parameters $ q $, $ \gamma_{1} $, and $ \gamma_{2} $ lead to different warning results. To show the influences of parameters $ q $, $ \gamma_{1} $, and $ \gamma_{2} $ on the warning performance, we use a warning frequency to describe the difference, given by:

\begin{subequations}\label{eq4.4}
	\begin{align}
	\eta_{1} & = \frac{N_{TLC-PDM}}{N} \\
	\eta_{2} & = \frac{N_{TLC}}{N} \\
	\eta_{3} & = \frac{N_{TLC-DSPLS}}{N} 
	\end{align}
\end{subequations}
where $ N_{TLC-PDM} $ and $ N_{TLC} $ are the number of warning points using the TLC-PDM method and the basic TLC method, respectively; $ N $ is the total number of driving data. A larger value of $ \eta_{1} $, $ \eta_{2} $ or $ \eta_{3} $ indicates the corresponding method obtains a higher warning frequency.
%
%

According to criteria (\ref{eq4.1}a) -- (\ref{eq4.1}c), we know that $ N_{TLC-PDM} $ could not be larger than $ N_{TLC} $ (i.e., $ \eta_{1} \leq \eta_{2}$) for the same driver, because the number of warning data points is always less than the total number of driving data. Fig. \ref{fig5.4} gives an example of the influences of parameters $ q $, $ \gamma_{1} $, and $ \gamma_{2} $ on the warning performance $ \eta_{1} $ for driver \#10 and the warning performance with the basic TLC is $ \eta_{2} = 0.2034 $. The number of warning points increases with values of $ \gamma_{1} $ or $ \gamma_{2} $ increasing, but is always less than $ N_{TLC} $. For example, in Fig. \ref{fig5.4}, we can note that when $ (\gamma_{1}, \gamma_{2} ) = (0.6,0.6) $, $ \eta_{1} $ approaches the maximum value but is always less than $ \eta_{2} = 0.2034 $ for all $ q $. 

For a fixed $ q $, we know that a larger value of parameters $ \gamma_{1} $ or/and $ \gamma_{2} $ generates a higher warning frequency. Parameters $ \gamma_{1} $ and $ \gamma_{2} $ in the warning strategy (\ref{eq4.1}a) -- (\ref{eq4.1}c) allow designers to set a preferred parameter for drivers. For instance,  by setting a larger value of $ \gamma_{1} $, the warning frequency will be larger, but a larger $ \gamma_{1} $ can guarantee a higher safety level. Parameter $ \gamma_{1} $ has the same influence as parameter $ \gamma_{2} $ on warning frequency.

\begin{figure}[t]
	\centering
	\includegraphics[scale = 0.6]{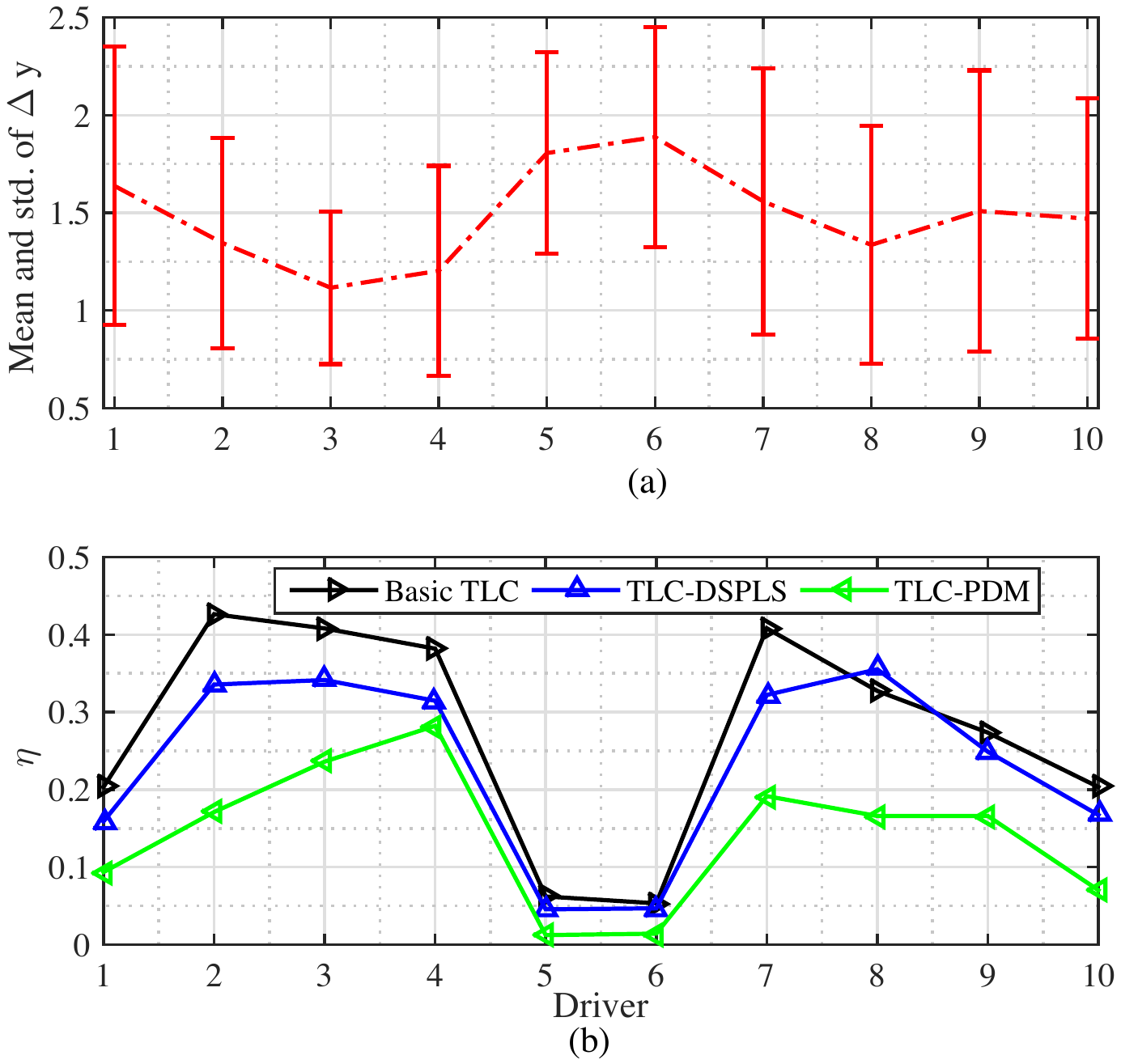}
	\caption{Relationship between warning frequency and driving styles.}
	\label{road}
\end{figure}

For the same value of parameters $ q $, $ \gamma_{1} $ and $ \gamma_{2} $, the warning frequency also differs between drivers. Table \ref{Tab2} lists the experimental results of warning frequency for 10 drivers with parameters $ \gamma_{0} =1 $, $ \gamma_{1} = -0.05 $, $ \gamma_{2} = 0.1 $, and $ q = 10 $. For driver \#5 and driver \#6 the TLC-PDM obtains a very lower warning frequency at about 0.013. Driver \#1 and driver \#10 can reach a middle level of warning frequency at about 0.08. For other drivers, however, the value of warning frequency is greater than 0.16. The difference across drivers is due to their differing driving styles. Drivers who prefer to drive a car in the middle of the road will naturally get a lower warning frequency; but some drivers, however, prefer to drive close to the road boundary. Fig. \ref{road} presents the relationship between driving styles and warning frequency. We know that driver \#5 and driver \#6, who usually drive a vehicle in the center of the road (i.e., $ \Delta y \approx 1.75 $ m, which is half the width of the driving lane), can get a lower warning frequency. Conversely, driver \#3 and driver \#4, who drive close to the boundary of the driving lane, get a higher warning frequency. 

\begin{table}[t]
	\centering
	\caption{Warning Frequency for Different Methods with $ \gamma_{0} = 1 $, $ \gamma_{1}  = -0.05 $, $ \gamma_{2}  = 0.1$, $ q = 10 $ and Prediction Time is 1 s.}
	\begin{tabular}{c|c|c|c}
		\hline
		\hline
		Driver & $ N_{TLC-PDM}(\eta_{1}) $ & $ N_{TLC}(\eta_{2}) $  & $ N_{TLC-DSPLS}(\eta_{3}) $  \\
		\hline
		1 & 0.0936 & 0.2034 & 0.1590 \\
		2 & 0.1711 & 0.4263 & 0.3355 \\
		3 & 0.2361 & 0.4079 & 0.3414 \\
		4 & 0.2823 & 0.3818 & 0.3146 \\
		5 & 0.0124 & 0.0622 & 0.0456 \\
		6 & 0.0143 & 0.0534 & 0.0369 \\
		7 & 0.1916 & 0.4081 & 0.3221 \\
		8 & 0.1660 & 0.3273 & 0.3552 \\
		9 & 0.1660 & 0.2735 & 0.2489 \\
		10 & 0.0703 & 0.2034 & 0.1673 \\
		\hline
		Average & 0.1404  &  0.2747  &  0.2336 \\
		\hline
		\hline
	\end{tabular}
	\label{Tab2}
\end{table}

\subsubsection{False-Alarm Rate}

\begin{figure}[t]
	\centering
	\includegraphics[scale = 0.65]{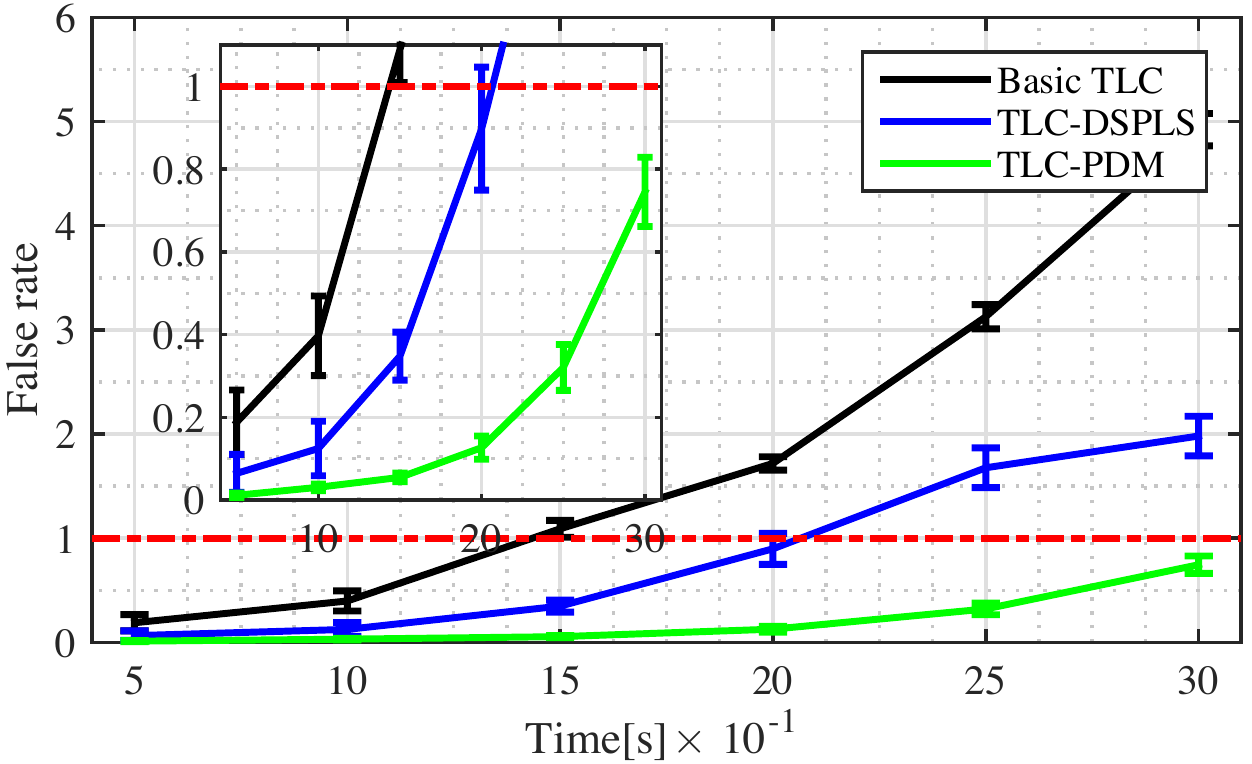}
	\caption{Comparison of the FAR between the basic TLC method, TLC-DSPLS method, and TLC-PDM method. For the TLC-DSPLS method, we set $ \gamma_{0} =1$. For the TLC-PDM method, we set $ \gamma_{1} = -0.05 $ and $ \gamma_{2} = 0.1$. }
	\label{fig5.5}
\end{figure}

We assess the performance of the LDW system using an FAR. The false warning event is recorded as a warning is triggered when the DCB occurs during the upcoming prediction interval. The FAR is computed by 
%
%

\begin{equation}\label{eq4.5}
\lambda = \frac{\# \ \mathrm{of \ false \ warning \ event}}{\# \ \mathrm{of \ all \ warning \ event}}
\end{equation}
For example, when the prediction time is 1 s, if a warning is triggered at time $ t $ and the DCB will occur during $ [t,t+1] $, then the warning is a false warning. For the TLC-DSPLS method, the DCB behavior is defined by the slope of a piecewise lateral displacement. See \cite{angkititrakul2011use}. For the TLC-PDM method, the DCB is determined by parameter $ \gamma_{2} $. 

To further show the benefits of the TLC-PDM method, we defined that  the DCB occurred when the future lateral displacement $ \Delta y_{t+q\Delta t} $ was larger than 0.1. Fig. \ref{fig5.5} shows the experimental results of FAR for three different methods, where $ \gamma_{1} = -0.05 $. The average and standard variance values  of FAR for driver \#10 are recorded. The FAR for LDWS with the basic TLC method was from 18.8 \% and 492.2 \%. The  FAR for the LDW system with the TLC-DSPLS method was from 6.44\% and 198.1\%. The FAR for the proposed TLC-PDM method was between 1.13 \% and 74.51 \%. 

For the basic TLC method, the FAR will reach over 100 \% when the prediction time is 1.5 s. For the TLC-DSPLS method, the FAR nearly reaches 100 \% when the prediction time is 2.0 s because the average error of LCB and DCB classification increases fast  with the increase in prediction time \cite{angkititrakul2011use}. For the TLC-PDM method, the largest FAR is about 75 \%, as the proposed prediction algorithm can predict the future lateral trajectory and its tendency. 


\section{Further Discussion and Future Work}
We proposed a TLC-PDM method for the LDW system, reducing the FAR caused by driver correction behavior. We also developed a model-based prediction algorithm for predicting vehicle trajectory and as a warning strategy for an LDW system. The experimental results show good performance for predicting the lateral trajectories of a vehicle. In this paper, we mainly focus on the framework of methods and the experiment validation in a driving scenario with small road curvature.

\subsection{Parameter Selection}
In this research, the parameters ($ \gamma_{1} $, $ \gamma_{2} $, $ \tau $, and $ q $) in the warning strategy  were defined subjectively by the researchers. We should note that these parameters can be tuned and set according to different design requirements. For example, if the designers wish to pay more attention to the safety level than to the warning frequency, they can set a larger value of $ \gamma_{1} $ or/and $ \gamma_{2} $. Furthermore, from Fig. \ref{road}, we know that driving styles are highly related to warning performance.  In future work, driver characteristics will be considered based on the TLC-PDM method to design a more driver-friendly LDW system.

\subsection{Vehicle Dynamics and Road Curvature}
In the model-based prediction algorithm  (\textbf{Algorithm 1}), the attributes of vehicle dynamics are not considered. Lateral trajectory prediction using vehicle dynamics (e.g., steering angle, brake/gas pedal position, acceleration) can improve the prediction accuracy \cite{mammar2006time}. In our work, we treated vehicle speed and road curvature as a constant during the iteration process and did not consider the effects of steering angle. Furthermore, road profiles (such as straight road or curved road with or without constant road curvature) greatly impact warning functions, see \cite{mammar2006time}. We use a simple TLC-based risk function to determine whether a warning should be given. In future work, we will take vehicle dynamics and more complicated road profiles into consideration in order to improve the performance of the TLC-PDM method.
%
%

\subsection{Driver State and Intention}
We do not classify or recognize the drivers' physical and physiological states \cite{kochhar2016robust}, such as the level of drowsiness or fatigue \cite{forsman2013efficient,saito2016driver}, drunkenness, or, aggression\cite{wang2016rapid}. Recognizing drivers' intentions, such as lane changing, is also beyond the scope in this paper. To design a more acceptable LDW system for drivers, drivers' intentions should be considered. Therefore, in future applications of the TLC-PDM method,  drivers' lane change behaviors \cite{sivaraman2014dynamic,libayesian} must be adequately recognized and considered. 

\section{Conclusion}
In this paper, a new method (TLC-PDM method) for reducing the false-warning rate of lane-departure warning systems is proposed. First, we model driver behaviors using five variables, including vehicle speed, relative yaw angle, the change rate of the relative yaw angle, road curvature, and relative lateral displacement. Then a personalized driver model is developed by combining the Gaussian mixture model and the hidden Markov model. The model parameters are determined by learning from  the naturalistic driving data collected from 10 drivers. Based on the personalized driver model, a model-based prediction algorithm for predicting the upcoming lateral vehicle trajectory is proposed and validated. Second, based on personalized driver model and the proposed prediction algorithm, a warning strategy is also developed. We also discuss the influences of different parameters of warning strategy on the warning performance. Last, to show the advantages of the TLC-PDM method, we compare it with the basic TLC method and the TLC-DSPLS method. The experimental results show that the TLC-PDM method can predict the upcoming lateral trajectory of a vehicle and obtain the lowest false-warning rate of 3 \% with 1 s prediction time.

\appendices
\section{}
In this Appendix, the calculation method of transfer matrix, $ \mathcal{T} $, in Section III, {\it C. Hidden Markov Model} is presented. 
Given the training data set with $ n $ data points $ \boldsymbol{\xi}_{t} $:
\begin{equation*}
\mathcal{S}_{Train} = \{ \boldsymbol{\xi}_{1}, \boldsymbol{\xi}_{2}, \cdots, \boldsymbol{\xi}_{t}, \cdots, \boldsymbol{\xi}_{n} \}
\end{equation*}
For each point $ \boldsymbol{\xi}_{t}  $, we define $ \boldsymbol{\xi}_{t}  $ is subject to mode $ m_{k} \in \mathcal{M} = \{1, 2, \cdots, K\} $ if 

\begin{equation}
m_{k} = \underset{k \in \{ 1,2, \cdots, K\} }{\max}  \ \mathcal{N}(\boldsymbol{\xi}_{t}; \boldsymbol{\mu}_{k}, \Sigma_{k})
\end{equation}

Therefore, each point $ \boldsymbol{\xi}_{t}  $ has a mode $ m_{t} \in \{ 1, 2, \cdots, K \} $, and we have a mode sequence with the same number of training data points
\begin{equation}\label{appedix.1}
\{m_{t}\}_{t=1}^{n} \Longleftrightarrow \mathcal{S}_{Train} = \{\boldsymbol{\xi}_{t}\}_{t=1}^{n}   
\end{equation}
Based on (\ref{appedix.1}), we estimate the transfer probability between mode $ i $ and mode $ j $ by

\begin{equation}
\alpha_{i,j} = \frac{F_{i,j}}{n_{i}}, \ i, j = 1, 2, \cdots, K
\end{equation}
where $ F_{i,j} $ is the count of transferring from mode $ m_{i} $ to $ m_{j} $, and $ n_{i} $ is the total number of training data points at mode $ i $. In this paper,  $ \sum_{i=1}^{K} n_{i} = n > 5\times 10^{5} $. Finally, we can get the transfer matrix $ \mathcal{T} $ as

\begin{equation}
\mathcal{T} = 
\begin{bmatrix}
\alpha_{1,1} & \alpha_{1,2}  & \cdots & \alpha_{1,K-1} & \alpha_{1,K} \\ 
\alpha_{2,1} & \alpha_{2,2}  & \cdots & \alpha_{2,K-1} & \alpha_{2,K} \\ 
\vdots & \vdots   &  \ddots & \vdots  &  \vdots \\ 
\alpha_{K,1} & \alpha_{K,2}  & \cdots & \alpha_{K,K-1} & \alpha_{K,K}\\ 
\end{bmatrix}_{K\times K}
\end{equation}




\ifCLASSOPTIONcaptionsoff
  \newpage
\fi



\bibliographystyle{IEEEtran}
\bibliography{LD_bib}
\end{document}